\pdfoutput=1

\documentclass[11pt]{article}

\usepackage[preprint]{acl}

\usepackage{times}
\usepackage{latexsym}

\usepackage[T1]{fontenc}

\usepackage[utf8]{inputenc}

\usepackage{microtype}

\usepackage{inconsolata}
\usepackage{graphicx}
\usepackage{array}
\usepackage{amsmath}
\usepackage{amssymb}
\usepackage{adjustbox}
\usepackage{algorithm}
\usepackage{array, makecell}
\usepackage{booktabs}
\usepackage{colortbl}
\usepackage{float}
\usepackage{graphicx}

\usepackage{times}
\usepackage{latexsym}
\usepackage{import}
\usepackage{multirow}
\usepackage{paralist}

\usepackage{wrapfig}
\usepackage{tabularx}

\usepackage{xcolor}
\usepackage{xspace}

\definecolor{LightCyan}{rgb}{0.88,1,1}
\definecolor{graylight}{gray}{0.7}
\definecolor{graymedium}{gray}{0.85}
\definecolor{graydark}{gray}{0.95}

\colorlet{lightblue}{blue!15}
\colorlet{lightorange}{orange!15}
\colorlet{lightgreen}{green!15}

\newcommand{\draftonly}[1]{#1}
\renewcommand{\draftonly}[1]{}

\newcommand{\frames}{\texttt{OATH-Frames}\xspace}
\newcommand{\definitions}{\texttt{OATH-Definitions}\xspace}
\newcommand{\guidelines}{\texttt{OATH-Guidelines}\xspace}

\newcommand{\numExpAnn}{4.1K\xspace}
\newcommand{\numLLLMassistAnn}{4.1K\xspace}
\newcommand{\numModelAnn}{2.4M\xspace}

\newcommand{\critiquecolor}[1]{{\colorbox{lightblue}{#1}}}
\newcommand{\perceptioncolor}[1]{{\colorbox{lightgreen}{#1}}}
\newcommand{\responsecolor}[1]{{\colorbox{lightorange}{#1}}}

\newcommand{\critiques}{\texttt{\critiquecolor{Critiques}}\xspace}
\newcommand{\perceptions}{\texttt{\perceptioncolor{Perceptions}}\xspace}
\newcommand{\responses}{\texttt{\responsecolor{Responses}}\xspace}
\newcommand{\finer}{\texttt{Issue-specific}\xspace}
\newcommand{\filter}{\texttt{Public Opinion}\xspace}

\newcommand{\expert}{\texttt{Expert}\xspace}
\newcommand{\gptsolo}{\texttt{GPT-4}\xspace}
\newcommand{\gptintheloop}{\texttt{Expert+GPT-4}\xspace}

\newcommand{\flanxl}{\texttt{Flan-T5-XL}\xspace}
\newcommand{\flanlarge}{\texttt{Flan-T5-Large}\xspace}
\newcommand{\llama}{\texttt{Llama-2-7b-chat-hf}\xspace}
\newcommand{\flanbase}{\texttt{Flan-T5-Base}\xspace}
\newcommand{\flansmall}{\texttt{Flan-T5-Small}\xspace}
\newcommand{\flan}{\texttt{Flan-T5}\xspace}
\newcommand{\governmentcritique}{\texttt{\critiquecolor{GovCrit.}}\xspace}
\newcommand{\moneyaid}{\texttt{\critiquecolor{MoneyAid.}}\xspace}
\newcommand{\societalcritique}{\texttt{\critiquecolor{SocCrit.}}\xspace}
\newcommand{\solutionsinterventions}{\texttt{\responsecolor{SolnInt.}}\xspace}
\newcommand{\personalinteraction}{\texttt{\perceptioncolor{Interact.}}\xspace}
\newcommand{\mediaportrayal}{\texttt{\perceptioncolor{MediaPort.}}\xspace}
\newcommand{\nimby}{\texttt{\perceptioncolor{NIMBY}}\xspace}
\newcommand{\deservingundeserving}{\texttt{\perceptioncolor{(Un)Deserv.}}\xspace}
\newcommand{\harmfulgeneralization}{\texttt{\perceptioncolor{HarmGen.}}\xspace}

\DeclareSymbolFont{extraup}{U}{zavm}{m}{n}
\DeclareMathSymbol{\vardiamond}{\mathalpha}{extraup}{87}
\DeclareMathSymbol{\varheart}{\mathalpha}{extraup}{86}
\newcommand{\aspace}{\hspace{1em}}
\newcommand{\uscsw}{$^{\vardiamond}$}
\newcommand{\usc}{$^{\varheart}$}

\title{\frames: Characterizing Online Attitudes Towards Homelessness\\ with LLM Assistants}

\author{
    Jaspreet Ranjit\usc\aspace 
    Brihi Joshi\usc\aspace
    Rebecca Dorn\usc\aspace
    Laura Petry\uscsw\aspace\\
    \textbf{Olga Koumoundouros}\uscsw\aspace
    \textbf{Jayne Bottarini}\usc\aspace 
    \textbf{Peichen Liu}\uscsw\aspace\\
    \textbf{Eric Rice}\uscsw\aspace
    \textbf{Swabha Swayamdipta}\usc\\
    \usc Thomas Lord Dept. of Computer Science, University of Southern California \\
    \uscsw Suzanne-Dwork School of Social Work, University of Southern California \\
    \texttt{jranjit@usc.edu}
}

\begin{document}
\maketitle

\begin{abstract}

\textit{\textcolor{red}{Warning: Contents of this paper may be upsetting.}}
Public attitudes towards key societal issues, expressed on online media, are of immense value in policy and reform efforts, yet challenging to understand \textit{at scale}.
We study one such social issue: homelessness in the U.S., by leveraging the remarkable capabilities of large language models to assist social work experts in analyzing millions of posts from Twitter.
We introduce a framing typology: Online Attitudes Towards Homelessness (\textbf{\texttt{OATH}}) Frames: nine hierarchical frames capturing critiques, responses and perceptions.
We release annotations with varying degrees of assistance from language models, with immense benefits in scaling: 
6.5$\times$ speedup in annotation time while only incurring a 3 point F$_1$ reduction in performance with respect to the domain experts. 
Our experiments demonstrate the value of modeling \frames over existing sentiment and toxicity classifiers. 
Our large-scale analysis with predicted \frames on 2.4M posts on homelessness reveal key trends in attitudes across states, time periods and vulnerable populations, enabling new insights on the issue.
Our work provides a general framework to understand nuanced public attitudes at scale, on issues beyond homelessness.
\end{abstract}

\section{Introduction}
\label{sec-intro}

\begin{figure*}[ht!]
 \includegraphics[width=\textwidth]{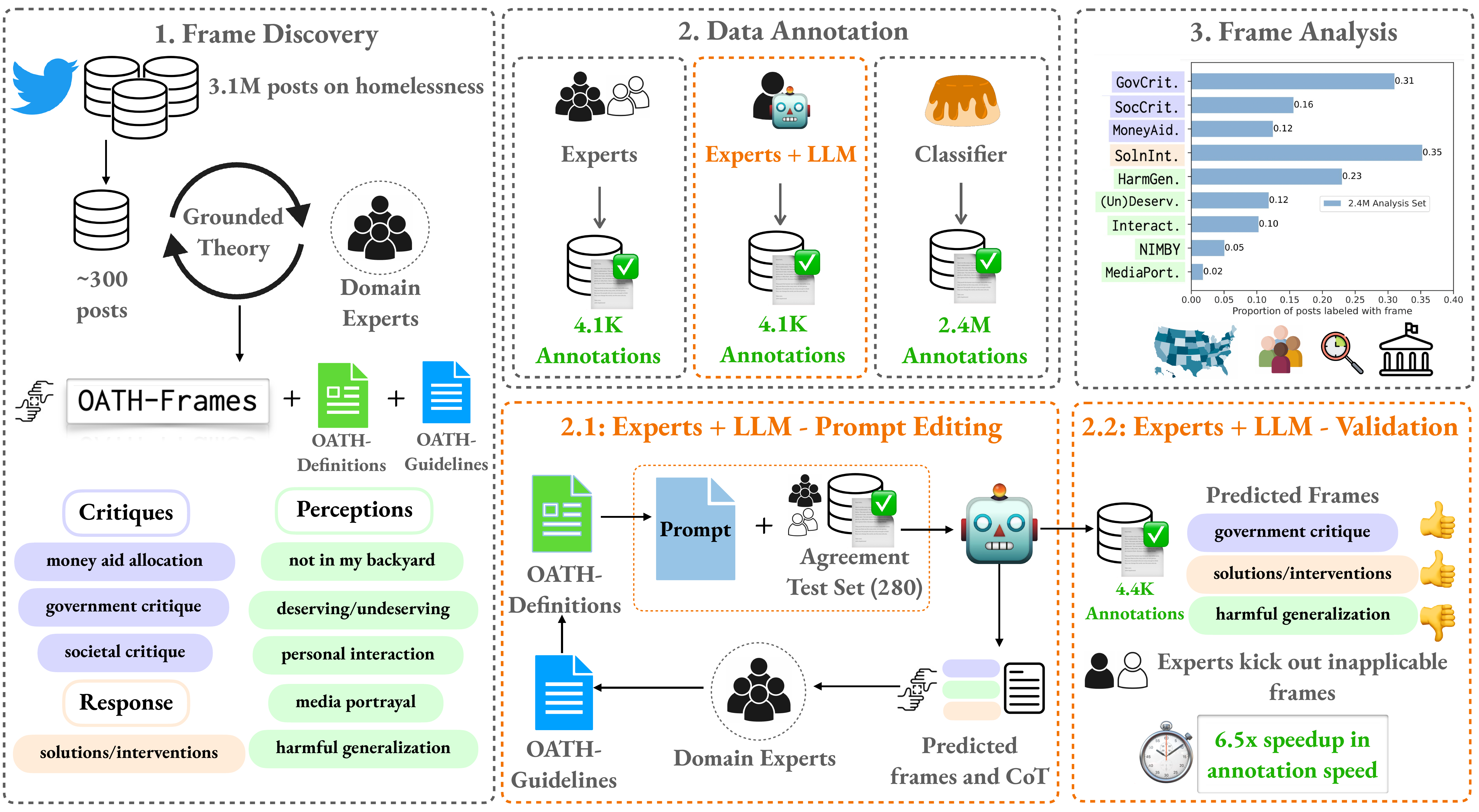}
 \caption{
 \textbf{Overview of \frames Collection.} (1.) Domain experts applied grounded theory to surface nine \finer frames, their corresponding definitions (\definitions), and annotation guidelines (\guidelines). (2.) We annotated posts with \frames via  Experts, via Experts + LLM (\texttt{GPT-4}), and via Multlilabel Classifier (\flanlarge). 
 Our Experts + LLM annotations pipeline consists of (2.1) prompt editing based on insights from domain experts and chains-of-thought, and (2.2) validation of predicted frames. 
 (3.) We analyzed annotated posts for attitudes across social and political dimensions. 
 }
 \label{fig:oath-frames}
\end{figure*}

Public opinion towards social issues is used as a signal by advocacy organizations to garner support for their relief and reform efforts \cite{burstein2020determinants}. 
Online social media platforms, such as Twitter\footnote{Now, X: \url{https://x.com}.}, serve as a great repository of public opinion from a diverse (e.g., socio-economic and political) user base \cite{hu2021socioeconomic}.
However, understanding broader public opinion necessitates processing online discourse \textit{at a large scale}, a daunting challenge in social science research.
Large language models (LLMs) have recently been used in social science \cite{ziems2023can,cho2023can,li2023coannotating, bohra2023byoc}, and may offer a promising solution to the scale challenge.
However, can LLMs be reliably used to process discourse with nuanced latent attitudes towards complex societal issues?

One such issue is homelessness in the U.S.\footnote{Over 653,104 U.S. individuals experienced homelessness in 2023 \cite{Sousa_Andrichik_Prestera_Rush_Tano_Wheele_2023}.
}, which inspires complex and nuanced public attitudes and beliefs \cite{Robertson1992}.
For instance, it is common for people to simultaneously feel both sympathy towards the plight of people experiencing homelessness (PEH), as well as anger and resentment towards them \cite{ cronley_2010, Moore_Sink_Hoban_Moore_1988, voices2002}.
A majority of work from social science has studied discourse about homelessness via ethnographic studies, qualitative surveys, and smaller scale content analyses \cite{coco_berkeley_homelessness, tsai2019public, kingree1997preliminary}. 
However, understanding the societal attitudes about homelessness \textit{at a large scale} via online discourse has rarely been attempted in prior efforts.
Our work investigates the role of LLMs in reliably assisting social work experts in characterizing such large-scale online discourse on homelessness.

In this work, we introduce a novel framework to understand, synthesize and characterize large-scale public attitudes towards societal issues through a collaboration between social work experts and LLMs, illustrated in \autoref{fig:oath-frames}.
Our framework consists of three components designed to study written discourse on social issues: (i) the discovery of a framing typology (\S\ref{sec:oath-frames}), (ii) annotation of data at scale with this typology (\S\ref{sec:oath-annotation}), and (iii) the analysis of these collected annotations (\S\ref{sec:oath-analysis}).
While we focus our study on homelessness, our framework is generalizable to other domains of social discourse.

At the heart of our framework is our novel framing typology to characterize Online Attitudes Towards Homelessness (\frames) in the U.S., as expressed in social media posts from Twitter (\S\ref{sec:oath-frames}). 
Our frames are developed using framing theory \cite{framing_entman} and grounded theory \cite{groundingtheory} from social science, and in collaboration with social work researchers studying homelessness. 
\frames can be broadly categorized into three themes: \critiques, \perceptions and \responses to PEH, and further into nine finer-grained \finer frames; Table \ref{tab:frames} illustrates our typology with examples. 
We provide three kinds of \frames annotations for posts from Twitter: \numExpAnn expert-only (\S\ref{sec:expert-annotation}), \numLLLMassistAnn LLM-assisted expert (\S\ref{sec:llm_collaboration}) and \numModelAnn predicted annotations from a multi-label classification model (\S\ref{sec:predicted-oath}).
Our LLM-assisted expert annotation for socially sensitive tasks follows a new scalable framework which incorporates expert insights (\guidelines) 
and chain-of-thought (CoT) explanations for designing better LLM prompts.
This results in a 6.5$\times$ annotation speedup, while incurring only a 3 point F$_1$ reduction in accuracy compared to experts---an attractive trade-off for our multi-label annotation task. 

\begin{figure}[h!]
 \includegraphics[width=\columnwidth]{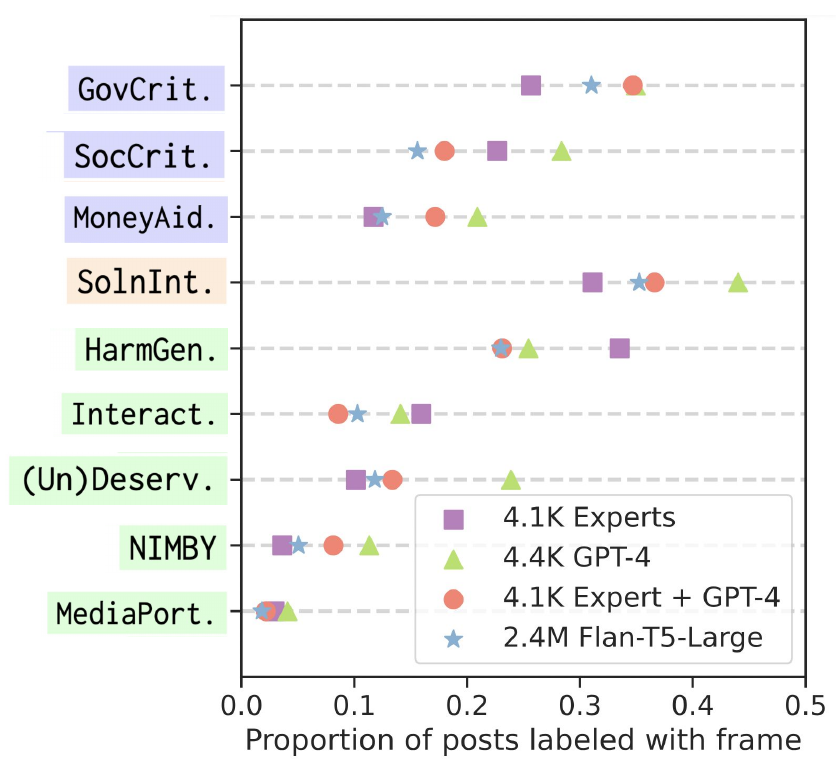}
 \caption{ \textbf{Proportion of \finer frames} across different annotation methods.
 Distribution of \frames across annotation strategies is similar. 
 See more details in \autoref{tab:appendix:annotation_splits} in  \autoref{sec:appendix:prediction}. 
 }
 \label{fig:human_prop}
\end{figure}

We observe that the most salient theme in expert-only annotations is \perceptions of PEH while the rest of our annotations highlight \critiques of the government, normative structures and (lack of) aid distribution (\autoref{fig:human_prop}). 

\begin{table*}[h!]
\centering
\resizebox{\linewidth}{!}{
\begin{tabular}{{ p{0.6\linewidth} p{0.99\linewidth} }}
    \toprule
      \textbf{Definition} & \textbf{Example and Labels from Different Annotators}\\ 
    \midrule
    \textbf{\governmentcritique} \textbf{Government Critique:} Criticism about government body, policies and laws including discussion of homelessness through the lens of political parties and values. & {\fontfamily{cmss}\selectfont these people are already housed possibly over crowded but at least housed .there are up to 130000 homeless living rough on our streets every night women and children in the back of vans and cars and the gov gives billions away to help third world countries.? time he looked here} \newline \textbf{\texttt{All}:} \governmentcritique \moneyaid, \deservingundeserving
    \\
    \midrule
    \textbf{\moneyaid} \textbf{Money Aid Resource Allocation:} Discussion of money, aid or resource disbursement on addressing the homelessness issue. Examples include the allocation of necessities such as essential items for emergency relief supplies, or government budgeting with respect to competing priorities. &  {\fontfamily{cmss}\selectfont you propose atlanta/fulton county fund a homeless city. i’m asking what return are we getting for our tax dollars? police officers and firefighters provide a service which can justify us building a training for them. what justification do you have for your city for the homeless?} \newline 
    \texttt{\textbf{All:}} \moneyaid, \governmentcritique / 
    \textbf{\expert, \gptsolo, \gptintheloop:}  \solutionsinterventions / \textbf{\expert:} \deservingundeserving 
    \\
    \midrule 
    \textbf{\societalcritique}  \textbf{Societal Critique:}  Criticism of social norms, systems and society at large in how homelessness is being addressed and perceived. Also includes pointing out hypocrisy and performative activism. &  {\fontfamily{cmss}\selectfont i really hate people who have mango trees but don’t eat mangos so they just let them go to waste. i be walking by peoples yards and see hella mangos just laying there for days. give them to the homeless, don’t just let the mangos sit there and rot.} \newline 
    \texttt{\textbf{All:}} \societalcritique, \solutionsinterventions
    \\
    \midrule 
    \textbf{\solutionsinterventions} \textbf{Solutions and Interventions:}  Discussion of solutions, interventions, charitable acts and remedies to address the homelessness crisis. & {\fontfamily{cmss}\selectfont Ever think that instead of jail, A MENTALLY ILL HOMELESS PERSON needs actual help? Yes, I have compassion for his victims. But I also have compassion for this guy. He needs help. He doesn’t need to be thrown in jail where he’ll rot and probably be left to die. You pointed out} \newline 
    \textbf{\texttt{All:}} \solutionsinterventions / 
    \textbf{\expert:}  \harmfulgeneralization / \textbf{\gptsolo:} \governmentcritique, \personalinteraction
    \\
    \midrule 
    \textbf{\nimby} \textbf{Not in my Backyard:}  Opposition by residents to proposed developments in their local area, as well as support for strict land use regulations against wanting to see homelessness in their local area and neighborhood.  & {\fontfamily{cmss}\selectfont just found out the head of the opposition to a local affordable housing for the homeless project is a senior planner in the neighboring county} \newline \textbf{\expert,\ \gptsolo,\ \gptintheloop:} \nimby, \governmentcritique / \textbf{\expert,\ \flanlarge:} \solutionsinterventions
    \\
    \midrule 
    \textbf{\personalinteraction} \textbf{Personal Interaction:}  Anecdote describing a direct personal exchange with PEH. & {\fontfamily{cmss}\selectfont i’m legit sad af. i saw a homeless woman and her daughter. i really wanted to help but i had no cash. i hope they are still there when i circle back around.} \newline
    \textbf{\texttt{All:}} \personalinteraction /
    \textbf{\expert,\ \gptsolo,\ \gptintheloop:} \solutionsinterventions 
    \\
    \midrule 
    \textbf{\mediaportrayal} \textbf{Media Portrayal:}  Portrayal of (fictional or real) PEH as described in the media (e.g. discussing PEH in a TV show or in the news). & {\fontfamily{cmss}\selectfont is the national news reporting on the homeless on the streets and encampments as well as random crime exaggerated? beautiful city and location.} \newline 
    \textbf{\texttt{All:}} \mediaportrayal / \textbf{\expert:} \nimby 
    \\
    \midrule 
    \textbf{\deservingundeserving}  \textbf{Deserving vs. Undeserving of Resources:}  Perpetuating a hierarchy of PEH with other marginalized communities or the use of harmful generalizations such as substance use and mental illness to justify that PEH that are more or less deserving of aid. Includes nationalistic rthetoric. & {\fontfamily{cmss}\selectfont Look at the illegals 500\$ a night hotel rooms , destroying them,  wasting food instead of giving to the homeless} \newline 
    \textbf{\texttt{All:}} \deservingundeserving, \moneyaid / 
    \textbf{\expert:} \solutionsinterventions / \textbf{\flanlarge: } \governmentcritique 
    \\
    \midrule 
    \textbf{\harmfulgeneralization} \textbf{Harmful Generalization:}  Blanket statements that ascribe an undesirable characteristic to PEH that include but are not limited to generalizing all PEH as having an unkempt appearance, or being violent, racist, thieves, or sexual predators. &  {\fontfamily{cmss}\selectfont @mention you're more confused than a homeless man in a house arrest. you can't even put your words together. please get out.} \newline \textbf{\texttt{All:}} \harmfulgeneralization, \societalcritique
    \\
    \bottomrule 
\end{tabular}
}
\caption{\textbf{\frames}: Definitions and examples for our 9 \finer \frames, accompanied by our \expert labels and \gptsolo, \gptintheloop, and \flanlarge predictions.
Each post is associated with multiple frames, at times even across our three overarching themes: (i)\critiques, (ii)\perceptions, and (iii)\responses.
}
\label{tab:frames}
\end{table*}

Our 2.4M \frames-annotated posts enable a large-scale analysis across states and time periods, revealing changing trends in attitudes with key sociopolitical events (\S\ref{sec:oath-analysis}).
\frames surface harmful language towards PEH, which we show are often mislabeled by popular sentiment and toxicity classifiers, highlighting the value of our typology.
We empirically establish trends where differences in attitudes are observed, when vulnerable populations
are compared with PEH; for instance, 
comparisons with immigrants elicit harmful rhetoric whereas those with Ukrainians elicit critiques towards the government for funding allocation. 
\frames can assist in understanding nuanced issues concerning marginalized PEH populations, often ignored in language technologies \cite{kiritchenko-etal-2023-aporophobia}.
Overall, our framework provides a blueprint for fast and scalable analysis of public opinion on key social issues of our time; we believe this will aid advocacy groups and political constituents in work towards meaningful social reforms \cite{burstein2020determinants, applebaum2001influence}. 
Our data and code will be publicly available\footnote{\url{https://dill-lab.github.io/oath-frames/}}.

\section{\frames Discovery}
\label{sec:oath-frames}

We discovered a novel typology of pragmatic frames for capturing and characterizing societal attitudes towards homelessness and people experiencing homelessness (PEH).
To this end, we applied concepts from framing theory \cite{framing_entman}, which has been used to identify how certain aspects in an issue's portrayal are emphasized to influence broader public opinion for topics like immigration \cite{mendelsohn2021modeling}, politics in media \cite{Boydstun2013IdentifyingMF}, eating disorders \cite{ArsenievKoehler2016ProanaPD},  and so on.
In social media discourse, prior work has focused on the framing of stigma towards PEH \cite{coco_berkeley_homelessness} and surveys that study attitudes and perceptions towards PEH \cite{attitudes_general_public_survey199, attitudes_robert_2015}.
In our work, we drew inspiration from \citet{coco_berkeley_homelessness} and their application of grounded theory \cite{groundingtheory} 
to characterize societal attitudes on homelessness \textbf{at scale}, going beyond stigma.
Following \citet{Boydstun2013IdentifyingMF}, our \textit{Frame Discovery} process uncovered salient themes (frame types) in a text corpus on a topic; in the context of our study, this resulted in \frames (Frames for \textbf{O}nline \textbf{A}ttitudes \textbf{T}owards \textbf{H}omelessness).

Using a keyword search for ``\texttt{homeless}'' on the {Twitter API},\footnote{\url{https://developer.X.com/en/docs/X-api} accessed in February 2023.} 
we sample 5M posts.
We preprocessed this sample (i.e. deduplicated, replaced emojis with unicode, anonymized user tags, and removed non-English posts) 
to retain 3.1M posts from 2.1M unique users spanning from 2021 to 2023. 

As a first step in \textit{Frame Discovery}, we only considered posts discussing public attitudes and opinions on homelessness, following \citeposs{Boydstun2013IdentifyingMF} \filter frame.
We discarded posts which we deemed ambiguous, under-specified or unrelated (e.g. political homelessness).
We also discarded posts containing first person lived experiences of PEH, since our focus was on studying public attitudes on homelessness.

Next, we followed an extensive iterative `coding' process based on grounded theory from social science \cite{groundingtheory} to surface our \frames where `codes' are 1-2 word descriptors of the data. 
The coding process involved our four domain experts independently analyzing a sample of 100 posts to conceptualize a unique categorization of attitudes. 
This was followed by a discussion where all posts were reviewed and similar frames were grouped together till we reached consensus on definitions, edge cases and representative examples. 
We repeated this process thrice on different sets of posts with refined iterations of our frames until we reached thematic saturation which surfaced nine fine-grained \finer frames under homelessness; see our \definitions in \autoref{tab:frames}.
Appendix \ref{sec:appendix:reminders} provides our experts' \guidelines developed during \textit{Frame Discovery} that contain detailed definitions and steps on how to address edge cases. 

Our \finer frames can be categorized into three overarching themes, as illustrated in \autoref{fig:oath-frames} (\textit{Frame Discovery}). 
First, we find \critiques of systems and institutions enabling homelessness and also the use of homelessness as a vehicle to critique others (e.g. actors, programs or spending). 
Second, we find posts reflecting different \perceptions of PEH which attribute undesirable characteristics to PEH to make harmful generalizations.
Third, we find many \responses to the issue, posed as solutions, interventions, calls to action and charitable acts.
Our three themes are consistent with those in prior work on homelessness \cite{inbook,gowan2010hobos}, but our work is the first to surface nine \finer (homelessness) frames providing more fine-grained insights.
\section{Annotation of \frames}
\label{sec:oath-annotation}
\begin{figure*}[h!]
 \includegraphics[width=\textwidth]{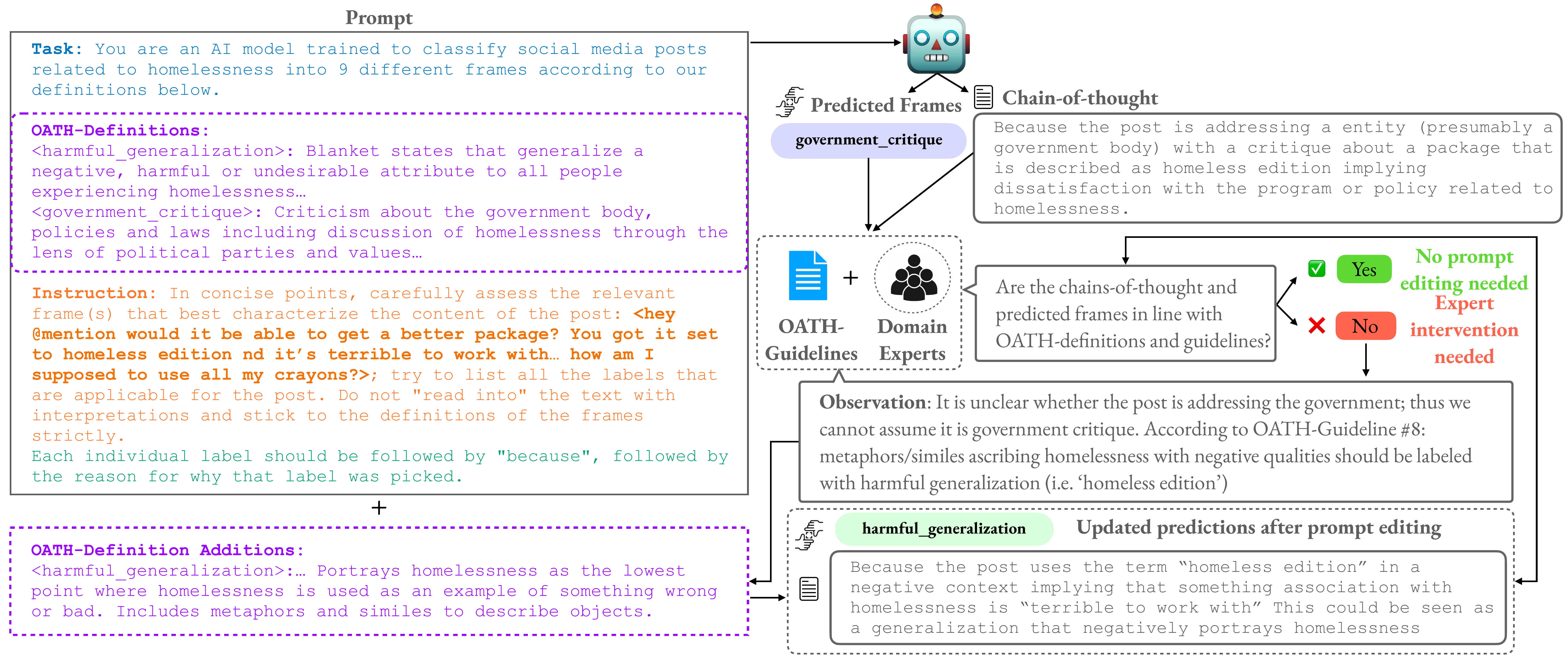}
 \caption{
 \textbf{Experts+LLM-Prompt Editing:} 
 We prompt \gptsolo with our task, \definitions, and our instruction as well as our own \guidelines and \gptsolo's Chain-of-Thought reasoning \citep[CoT;][]{wei2022chain}, in an iterative setup. We observe that CoT serves as a signal to understanding misinterpretations of \definitions and benefits from addition of expert annotator guidelines (\guidelines).
 }
 \label{fig:gpt_in_the_loop}
\end{figure*}

\subsection{Expert Annotation}
\label{sec:expert-annotation} 

Our four domain experts leading the frame discovery effort (\S\ref{sec:oath-frames}) trained two other annotators (trained experts). 
The trainees first familiarized themselves with \definitions and \guidelines as detailed in Appendix \ref{sec:appendix:reminders}. 
Next, the trainees labeled 20 posts with \frames where the experts clarified questions about \definitions. 
Finally, the trainees performed an independent annotation exercise on a set of 280 posts annotated by the domain experts. 
The training concluded once the trainee achieved moderate agreement (Fleiss' $\kappa$ > 0.4 \citealp[;][]{fleiss1971measuring}) with the experts. 

For all relevant posts, each expert (domain and trained) annotator performs a 9-way-multilabel classification of \finer frames, where 54\% of the posts were labeled with more than one frame. Co-occurrence of frames represented multiply expressed attitudes highlighting the complexity of our task.
All annotators achieved an average inter-annotator agreement of 0.46 (Fleiss' $\kappa$) and a macro F$_1$ of 0.55 (see \autoref{tab:res:finer_models_vs_everyone_f1}) across nine \finer frames.\footnote{Our agreement across 9 frames is `moderate'; prior work has reported similar agreement of 0.3-0.4 across 8 frames and 4 annotators in a multi-label classification task \cite{Stede_Bracke_Borec_Kinkel_Skeppstedt_2023}. We calculate the agreement across six annotators, a much higher number than prior work that considers only two annotators \cite{card2015media}.} 

Overall, we considered 5.3K posts, of which we filtered out 1.2K posts, leaving \numExpAnn expert-annotated posts. 
See Appendix \ref{sec:appendix:human_annotations} for details on our inter-annotator agreement, co-occurrence of frames and annotation interface. 

\subsection{LLM-Assisted Expert Annotation}
\label{sec:llm_collaboration}

\begin{table*}[ht]
\centering
\footnotesize
\begin{tabular}{lcccc|cc}
    \toprule
    \textbf{Annotators $\rightarrow$} & \textbf{\texttt{Experts$^*$}} & \textbf{\gptsolo} & \textbf{\gptintheloop} & \textbf{\flanlarge} & \textbf{\flanlarge}& \textbf{\texttt{Llama-2-7b}}\\ 
    \cmidrule(lr){2-7}
    \textbf{Test Set $\rightarrow$} & \multicolumn{4}{c|}{\textbf{$|D_\text{agreement}|=$280}} &  \multicolumn{2}{c}{\textbf{$|D_\text{test}|=$1280}}\\
    \midrule
    \governmentcritique 
     & 0.65 $\pm$ 0.06 & 0.70 $\pm$ 0.08 & \textbf{0.72 $\pm$ 0.07} & 0.65 $\pm$ 0.07 & 0.63& 0.35 \\
    \moneyaid
      & 0.60 $\pm$ 0.16 & 0.58 $\pm$ 0.14 & \textbf{0.61 $\pm$ 0.15} & 0.59 $\pm$ 0.18 & 0.61& 0.22 \\
     \societalcritique 
     & 0.43 $\pm$ 0.09 & 0.46 $\pm$ 0.06 & \textbf{0.50 $\pm$ 0.05} & 0.42 $\pm$ 0.09 & 0.49& 0.31\\
     \solutionsinterventions 
     & 0.69 $\pm$ 0.06 & 0.62 $\pm$ 0.06 & 0.65 $\pm$ 0.06 & \textbf{0.68 $\pm$ 0.03} & 0.57& 0.48\\
    \personalinteraction 
     & 0.45 $\pm$ 0.13 & 0.45 $\pm$ 0.07 & \textbf{0.52 $\pm$ 0.07} & 0.54 $\pm$ 0.11 & 0.70 & 0.18 \\
     \mediaportrayal 
     & \textbf{0.70 $\pm$ 0.10} & 0.50 $\pm$ 0.08 & 0.63 $\pm$ 0.11 & 0.69 $\pm$ 0.10 & 0.54 & 0.13\\
    \nimby 
     & 0.17 $\pm$ 0.21 & 0.30 $\pm$ 0.16 & \textbf{0.39 $\pm$ 0.21} & 0.34 $\pm$ 0.26 & 0.26 & 0.10\\
     \deservingundeserving 
     & 0.58 $\pm$ 0.19 & 0.46 $\pm$ 0.06 & 0.63 $\pm$ 0.08 & \textbf{0.67 $\pm$ 0.01} & 0.61 & 0.22\\
     \harmfulgeneralization 
     & \textbf{0.53 $\pm$ 0.07} & 0.49 $\pm$ 0.06 & 0.51 $\pm$ 0.06 & 0.55 $\pm$ 0.05 & 0.58 & 0.41\\
     \filter 
     & \textbf{0.75 $\pm$ 0.05} & 0.04 $\pm$ 0.05 & 0.04 $\pm$ 0.05 & 0.0 $\pm$ 0.0 & 0.0 & 0.02\\
     \midrule
     Aggregated$_{\text{Micro}}$
     & \textbf{0.62 $\pm$ 0.01} & 0.54 $\pm$ 0.05 & 0.59 $\pm$ 0.05 & 0.57 $\pm$ 0.04 & 0.55 & 0.30\\
     Aggregated$_{\text{Macro}}$
     & \textbf{0.55 $\pm$ 0.04} & 0.46 $\pm$ 0.04 & 0.52 $\pm$ 0.04 & 0.51 $\pm$ 0.03 & 0.50 & 0.24\\
    \bottomrule 
\end{tabular}

\caption{{Mean ($\pm$ s.d. across 6 expert annotator labels) F$_1$ for \finer frames and the \filter filter, and overall macro and micro F$_1$ (Aggregated)}.
\expert* reports the average F$_1$ computed using each annotator as a reference. 
\gptintheloop and \flanlarge consistently outperform \gptsolo, while lagging behind \expert*.
Performance of \flanlarge and \llama (zero-shot) are reported on  1280 singly annotated test instances. 
See detailed precision and recall for each row in Appendix \autoref{tab:res:finer_models_vs_everyone}. 
}
\label{tab:res:finer_models_vs_everyone_f1}
\end{table*}

Given the nature of the rhetoric on homelessness, annotating 5.3K posts was laborious, and emotionally challenging for the human experts. 
LLMs are now being deployed as annotation assistants in social science \cite{li2023coannotating, bohra2023byoc}, but questions on how best to incorporate insights from social science into LLMs still remain \cite{nelson2024designing, gondimalla2024labeling}; also see Appendix \ref{sec:related-llm-assistant}.
We explore a collaborative setting for annotating \frames with LLM assistants to our expert annotators to ameliorate their manual and emotional labor.
We first investigate the ability of \gptsolo \cite{achiam2023gpt} to do our annotation task \textit{stand-alone} (\S\ref{sec:gpt-only-annotations}), before describing our collaborative framework (\gptintheloop; \S\ref{sec:gpt-expert-annotations}).

\subsubsection{Predicting \frames with only \gptsolo} 
\label{sec:gpt-only-annotations}

We design our prompts for \gptsolo by incorporating instructions and examples as well as our own \guidelines and \gptsolo's own Chain-of-Thought reasoning \citep[CoT;][]{wei2022chain}, in an iterative setup as shown in \autoref{fig:gpt_in_the_loop}.
Building on prior work \cite{pangakis2023automated, rytting2023towards}, we observe that CoT serves as a signal to understanding misinterpretations of \definitions and benefits from addition of expert annotator guidelines (\guidelines).
Our prompt refinement pipeline could be generalized to other social science tasks, where practitioners could use LLM CoT and their expert annotator guidelines to leverage LLMs as annotation assistants.
Appendix \ref{sec:appendix:gpt_annotations} details the prompts we used. 
 
We report \gptsolo's performance on a subset of 280 multiply-annotated posts\footnote{This allows us to compute performance within a margin of error across all 6 (expert) annotators.} ($|D_\text{agreement}|=$280) in  \autoref{tab:res:finer_models_vs_everyone_f1}.
\gptsolo achieves a precision of 0.39 $\pm$ 0.05 and recall of 0.64 $\pm$ 0.05 across all nine \finer frames showing that it may have reasonable coverage of the predicted labels but also produces
false positives due to the inherent social nuances in the posts that may be easier for a expert to resolve. 
For example, \gptsolo often confuses the \societalcritique and \governmentcritique frames, seen in the following post: 
{
{\fontfamily{cmss}\selectfont`@mention it appears the problem in el paso with illegal aliens were solved overnight, so that you could have a photo op without anyone seeing hundreds of homeless illegals on the streets'}}.
Here, although the government is not explicitly mentioned, politicians often enact displacement sweeps and photo-ops to disguise the severity of the homelessness crisis. 
We observe that \gptsolo is not able to resolve domain-specific references (i.e. displacement sweeps) and thus predicts both \societalcritique and \governmentcritique (\autoref{tab:gpt4_error_analysis}). 
In these cases, it's important to have experts validate the predictions. 

\subsubsection{Predicting \frames with \gptintheloop} 
\label{sec:gpt-expert-annotations}

Although LLMs serve as a useful tool to efficiently generate annotations, the inherent subjectivity and sensitivity of our task poses a challenge to using \gptsolo as an annotator on its own \cite{rytting2023towards, heseltine2024large}. 
We propose an \gptintheloop approach that leverages both \gptsolo's annotation efficiency and human expertise to verify and correct \gptsolo's generations. 
We generate \gptsolo predictions on a set of 4.4K posts.
Through expert validation, we filter out 300 posts, resulting in  4.1K \gptintheloop validated posts. 
Our expert validation involves discarding false positive labels by \gptsolo; this results in a 12 point boost in precision and a slight drop in recall compared to \gptsolo generations. 
We note that by themselves, experts take 187.49 ($\pm$ 30.71) seconds, whereas with \gptintheloop, they take 28.80 ($\pm$ 8.55) seconds, which is a 6.5$\times$ boost in annotation speed.
Overall, \gptintheloop results in a 6 point increase in macro F$_1$ over \gptsolo; see Table \ref{tab:res:finer_models_vs_everyone_f1}. 

\subsection{Scaling \frames Annotations}
\label{sec:predicted-oath}
While \gptintheloop reduces manual labor, annotations by \expert and \gptintheloop are relatively cost-intensive at a larger scale. 
To this end, we scale \frames annotations using a model trained with \expert and \gptintheloop annotations.
We obtain 10.4k training examples: 5.3K \expert and 4.1K \gptintheloop annotations and 1K filtered posts. 
We split these into a train-val-test split containing 8217, 913 and 1280 posts respectively and train a \flanlarge \cite{flant5} model for 10-way multilabel classification (nine \frames and filter). 

Results in Table \ref{tab:res:finer_models_vs_everyone_f1} show that \flanlarge performs better than \gptsolo with a Macro F$_1$ of 0.51 $\pm$ 0.05, and is on par with the \gptintheloop setting on the Agreement Test Set ($|D_\text{agreement}|=$280). We also find that \flanlarge obtains a Macro F$_1$ score of 0.50 on a larger test set of 1280 posts ($|D_\text{test}|=$1280). 
Among four fine-tuned size-variants of \flan, and zero-shot \texttt{Llama-2-7b} \cite{touvron2023llama}, \flanlarge performs the best; detailed results and data settings are provided in Appendix \ref{sec:appendix:prediction}. 

Using our trained \flanlarge model, we generate \frames predictions on a set of 2.4M unlabeled posts. 
\autoref{fig:human_prop} illustrates that the distribution of frames under different annotation strategies is similar. 
Next, we use the 2.4M predicted annotations in a large-scale socio-political analysis of public attitudes.

\section{Analyzing Posts with \frames}
\label{sec:oath-analysis}

Synthesizing public opinion using online discourse is useful for advocacy groups and policymakers to inform homelessness services and equitable housing policies to local communities \cite{burstein2020determinants}.
We demonstrate the utility of \frames to such end users by analyzing attitudes with respect to state mentions, marginalized populations, and time period on the set of \numModelAnn posts.
We also highlight the utility of our \frames over sentiment and toxicity labels.

\subsection{\frames Versus Toxicity and Sentiment Prediction} 
\label{subsec:toxsent}
PEH experience harms online in unique ways through subtle stereotyping (i.e.\harmfulgeneralization: `\textit{you look homeless}'; \citealp{kiritchenko-etal-2023-aporophobia, hutchinson2020social}).
We investigate how existing toxicity and sentiment  
classifiers capture harmful rhetoric against PEH as represented by the \harmfulgeneralization frame. 
\begin{figure}[h!]
\centering
 \includegraphics[width=\columnwidth]{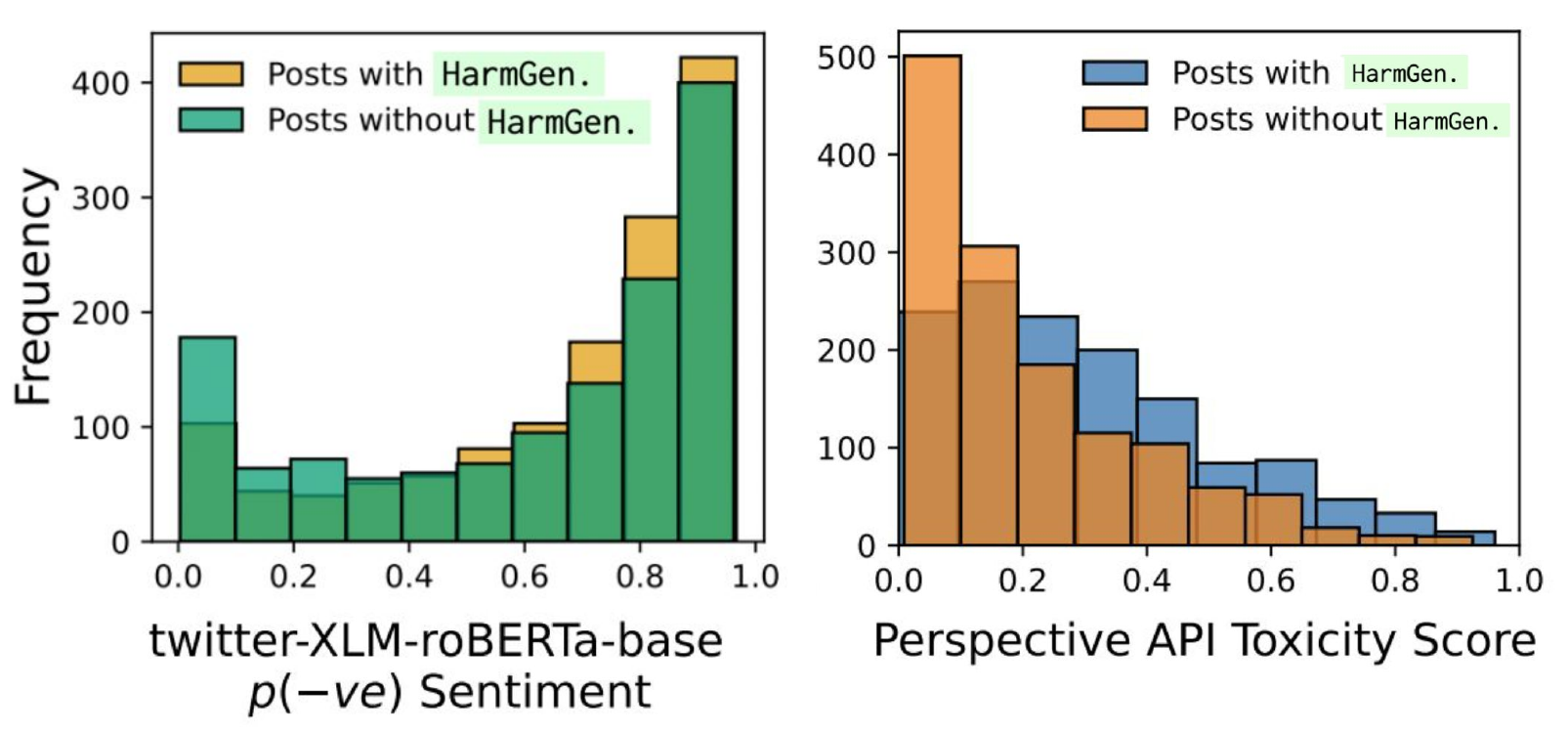}
 \caption{Distributions of sentiment and toxicity scores for posts labeled with and without \harmfulgeneralization in our subset of 4.1K \expert-annotated posts.}
 \label{fig:tox_sent_harmgen}
\end{figure}
We plot the distribution of the Perspective API\footnote{\url{https://perspectiveapi.com}} toxicity score and sentiment scores \cite{rosenthal2017semeval} in Figure \ref{fig:tox_sent_harmgen} for the subset of \expert-annotated posts labeled with \harmfulgeneralization.
We find that a majority of \harmfulgeneralization posts are labeled with a toxicity of less than 0.5, indicating a high false negative rate. 
For example, the following post contains harmful stereotypes against PEH but 
was tagged with a toxicity score of less than 0.5:
{\fontfamily{cmss}\selectfont `@mention some people wear clothes at home i wear for an outing and here i dress up like a homeless person inside my home whos a drug addict and has been sleeping since forever'}. 
Furthermore, the negative sentiment distributions for posts labeled with and without \harmfulgeneralization are significantly similar (\textit{independent samples t-test}: $p < .001$) indicating that sentiment classifiers are not able to represent \harmfulgeneralization against PEH.
This highlights that \frames provides a framework characterizing how PEH may experience harms and subtle stereotyping, going beyond sentiment and toxicity. 
Our framework can serve to improve existing classifiers in detecting such harms, especially for underrepresented communities \cite{dorn2023non}.

\subsubsection{Case Study: Solutions and Interventions in Posts Mentioning California}

Given the high rates of PEH in California (CA accounted for 49\% of U.S. unsheltered PEH in 2023), we investigate the proposed \solutionsinterventions for posts that mention CA. 
We measure the statistical prevalence of a term in CA posts in two subsets: posts labeled with \solutionsinterventions and posts labeled with a positive sentiment probability ($p(+ve)> 0.5$) \cite{rosenthal2017semeval} in our set of \numModelAnn posts. 
\begin{table}[h]
\centering
\small
\begin{tabular}{lclc}
\toprule
(+) \solutionsinterventions & Score & $p(+ve)$ & Score \\
\midrule
rent & 3.72  & great & 4.68 \\
voucher & 2.59  & love & 4.22 \\
abandon & 2.48  & new & 3.26 \\
bass & 4.68 & thank & 3.16 \\
healthcare & 2.03 & help & 2.32 \\
\bottomrule
\end{tabular}
\caption{Log-odds ratios for terms that are statistically more likely to appear in posts about CA labeled with (+) \solutionsinterventions and posts about CA with positive sentiment ($p(+ve) > 0.5$). 
Scores outside the range [1.96, -1.96] are considered statistically significant. Full set of results are included in \autoref{tab:logodds}. 
}
\label{tab:logodds_sub}
\end{table} 

We calculate \citeposs{Monroe_Colaresi_Quinn_2017} weighted log-odds ratio with a Dirichlet informed prior for unigrams/bigrams. 
By incorporating a prior estimate over the expected distribution of the vocabulary, this approach accounts for words that are very commonly or rarely prevalent in the two subsets of posts. 
\autoref{tab:logodds_sub} shows 5 significant terms in each subset and their z-score. 

We find that in posts labeled with (+) \solutionsinterventions, \frames captures more comprehensive insights specific to homelessness in CA. Whereas posts labeled with a positive sentiment ($p(+ve)$) contain more generic insights (i.e. `\textit{great, love, help}').
For example, `\textit{abandon, voucher}', and `\textit{bass}' refer to LA Mayor Karen Bass's homelessness initiative: `Inside Safe' that converted abandoned hotels to temporary shelters for PEH \cite{robinhotels2023} whereas `\textit{rent}' and `\textit{healthcare}' are considered as drivers of homelessness in CA and potential entry points for interventions. 
Although sentiment classifiers provide coarse-grained insights of conflating factors in relation to homelessness (e.g. substance use, crime), we find that \frames provide more targeted insights specific to CA.
\subsection{Attitudes across State Mentions}
\label{subsec:states}

\begin{figure}[h!]
 \includegraphics[width=\columnwidth]{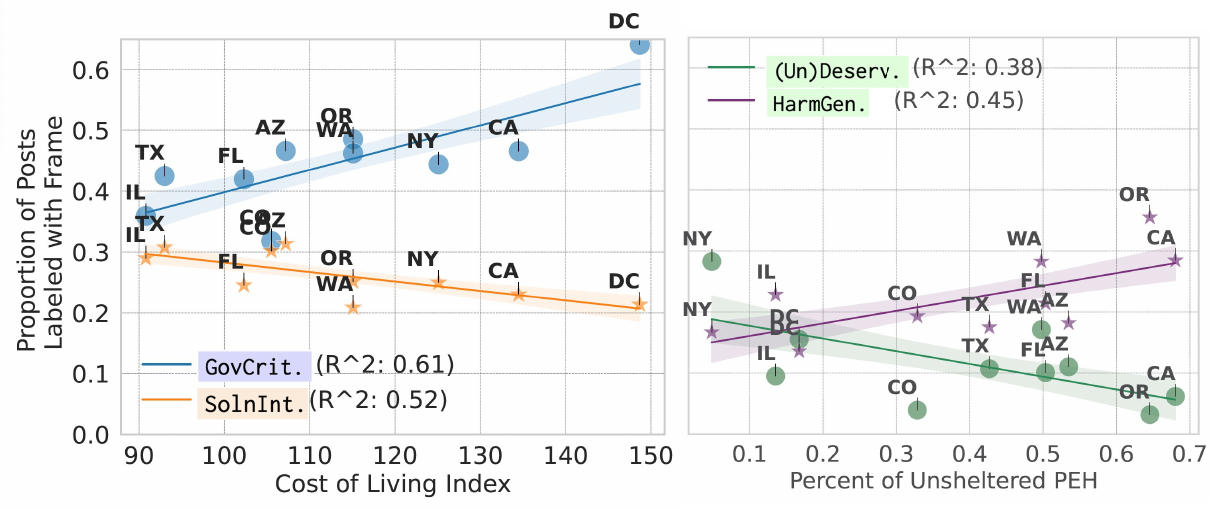}
 \caption{Linear regression results for state level factors: cost of living and percent of unsheltered PEH, predicting proportion of frames for each state. 
 The shaded region represents the standard error of the fit to the true values. 
 }
 \label{fig:regression}
\end{figure}

Discourse on homelessness is closely tied to a state's social and economic climate and the prevalence of PEH in each region \cite{tsai2019public}. 
We investigate the saliency of our \finer frames in posts that mention states in the U.S.\footnote{Our analysis includes the following states which were most frequently represented in our set of 12K posts containing state mentions out of 2.4M: CA, NY, TX, WA, OR, FL, IL, CO, AZ and DC across 12K posts.}
Given less than 1\% of tweets are geotagged \cite{doi:10.1177/0165551515602847}, we use NER\footnote{\url{https://huggingface.co/cjber/reddit-ner-place_names}} to segment our data by posts mentioning different U.S. states.
We find that \harmfulgeneralization is the most prominent \finer frame in posts about California, Washington and Oregon and \deservingundeserving and \moneyaid are the most prominent in posts about New York (see \autoref{fig:all_states}, \autoref{tab:state_examples} in Appendix \ref{sec-analysis}). 
To better understand the drivers of attitudes in a state, we also estimated a linear regression for two state level factors (i.e. cost of living index, percent of unsheltered PEH) in relation to the proportion of posts labeled with frames\footnote{Understanding the socio-economic factors shaping attitudes extends beyond cost of living and unsheltered populations \cite{lee1990causes}; we leave further exploration of these factors up to future work.} as shown in \autoref{fig:regression}. 
We find that a high cost of living index\footnote{Cost of living index indicates the amount needed to cover basic expenses, (i.e. food, shelter) \url{https://worldpopulationreview.com/state-rankings/cost-of-living-index-by-state}} correlates to a larger proportion of \governmentcritique ($R^2$ of 0.61) and fewer \solutionsinterventions whereas a larger unsheltered PEH population correlates to more \harmfulgeneralization ($R^2$ of 0.45) and less \deservingundeserving.
We hypothesize that greater visibility of PEH in the west coast, which is due to a larger unsheltered population (68\% in California compared to 4.9\% in New York), results in \harmfulgeneralization and in contrast, posts about New York are more likely to contain \deservingundeserving attitudes due to an influx of asylum seekers and immigrants in 2023 \cite{Sousa_Andrichik_Prestera_Rush_Tano_Wheele_2023}. 
In addition, higher costs of living elicit more \governmentcritique for economic reform and reducing rent prices. 

\subsection{Differences in Attitudes Across Vulnerable Populations and PEH}
Attitudes of \deservingundeserving \cite{solas2018deserving} reflect comparisons between vulnerable populations and PEH as being more or less deserving of aid \cite{kiritchenko-etal-2023-aporophobia, Hoffmann2019WhereFF} putting them at risk for being further ostracized from garnering public support for relief and reform measures \cite{lalor-etal-2022-benchmarking}. 
We investigate the finer-grained differences in attitudes as characterized by \frames when PEH are compared to other vulnerable populations in our subset of posts labeled with \deservingundeserving \cite{rosenheck1999special, Chishti_2022}.

We identify the following subpopulations of PEH:
\textit{veterans, people,  immigrants, refugees, americans} and the following vulnerable populations: \textit{asylum seekers, immigrants, refugees, illegal aliens, Ukrainians} amongst the most frequently occurring bigrams in our data. 
We extract a subset of 153K posts (out of 2.4M posts) mentioning these terms.
`Homeless Veteran', `Homeless People' and `Homeless American' co-occur most frequently with `Ukrainian' and `Immigrant' (see \autoref{fig:co_occur_identity} in Appendix \ref{sec-analysis}). 
For each co-occurring (PEH, Vulnerable Population) pair, we examine the most statistically significant attitudes (see \autoref{tab:co_occur_identity} in Appendix \ref{sec-analysis}). 
We find marked differences in attitudes when PEH are compared to Ukrainians versus when they are compared to immigrants. 
Public attitudes in posts that compare PEH with Ukrainians appear either as \governmentcritique or appeals for \moneyaid. 
In contrast, posts comparing PEH to immigrants elicit more nationalistic sentiment and are oriented around relocation (\nimby) and \harmfulgeneralization. 

\subsection{Tracking Attitudes Across Time}
\label{subsec:analysis-time}
\begin{figure}[t!]
 \includegraphics[width=\columnwidth]
 {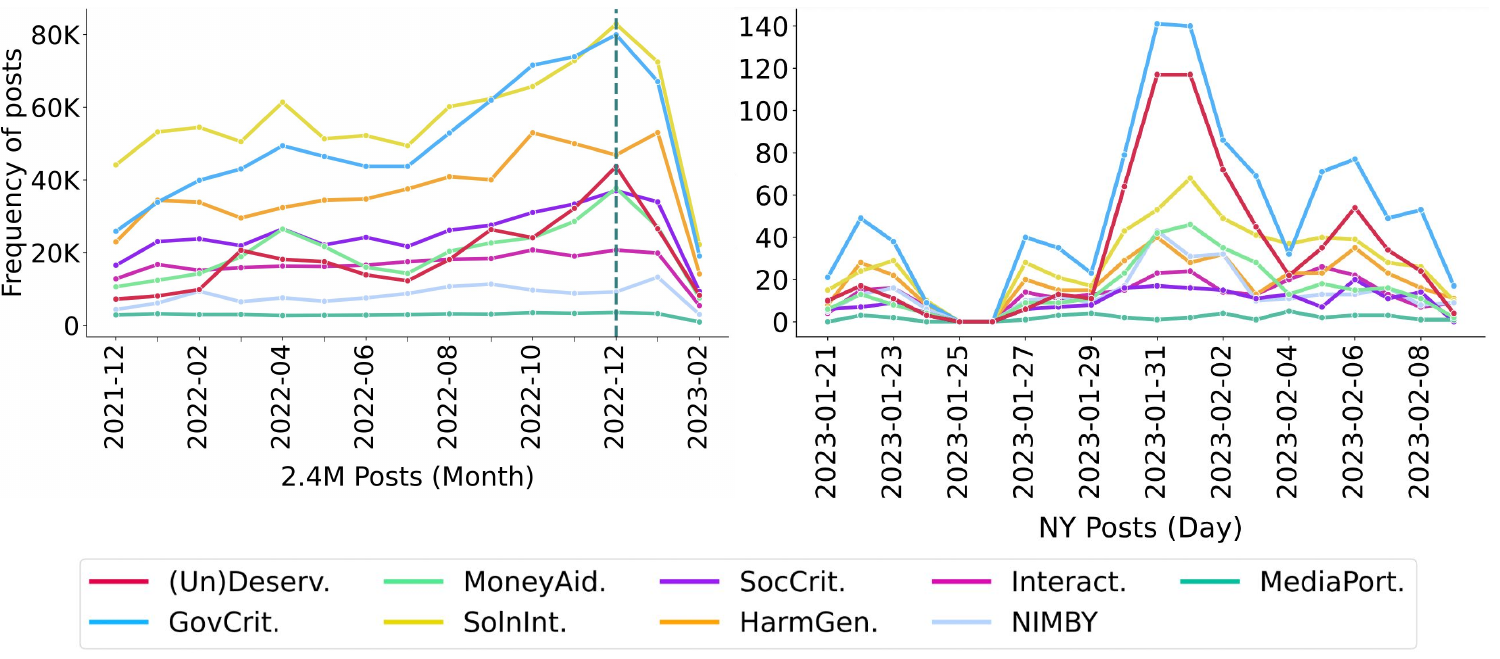}
 \caption{Frequency of posts labeled with frames across months in 2021-2023 in 2.4M annotated posts (Left) and across days in posts about NY (Right).}
 \label{fig:time}
\end{figure}

Geo-political events serve as catalysts for reactive and inflammatory attitudes on social media. 
We find that the prevalence of certain attitudes across several months correlates with news cycles that dominated the discourse on homelessness. 
Figure \ref{fig:time} shows the frequency of attitudes by month 
across 2021 to 2023. 
We observe that \governmentcritique, \solutionsinterventions, \deservingundeserving, and \moneyaid peaked 
in 12-2022, a time when tensions were rising in the U.S. due to Congress considering a spending bill with \$44.9 billion in aid for Ukrainians during the Russian-Ukraine War \cite{weber2022}. 
This sparked comments regarding the prioritization of government funding (\governmentcritique, \moneyaid) amongst Ukraine and PEH (\deservingundeserving). 

For example, the following post contains mentions of the spending bill followed by a comparison to the existing homelessness crisis in the country:  {\fontfamily{cmss}\selectfont `They just approved \$45 Billion for Ukraine that'll be unaccounted for and to be spent however they please. We have homeless wandering the streets looking for a warm place to get out of the cold'}. 

Additionally, we investigate the prevalence of attitudes in posts about New York and find that \governmentcritique and \deservingundeserving attitudes spike significantly between 01-29-2022 and 02-02-2022. 
The spike correlates to a \$40M contract to convert the hotels into temporary housing for an influx of migrants \cite{Zraick_2023, flanagan2023} leading to protests, increased hostility towards migrants and public opposition against the government. 
Such comparisons between immigrants and the state's
homeless population resulted in a spike in \deservingundeserving and \governmentcritique attitudes.
\section{Conclusion and Future Work}
\label{sec-conclusion}
We introduce \frames which characterize online attitudes towards homelessness under three overarching themes and nine \finer frames.
We annotate posts using \expert, \gptsolo and \gptintheloop and use them to train a model to predict
\frames on a large collection of 2.4M posts.
We show that modeling \frames offers
distinct advantages over existing models for toxicity and sentiment classification, and
reveals nuanced attitudes
in relation to states, marginalized populations, and over time. 

We hope to expand \frames to include more context such as target groups, or 
correlated factors (e.g., substance use and mental illness). 
Our work serves as a blueprint for building frameworks that incorporate insights from social science experts for different large-scale socially sensitive annotation tasks.
We expect \frames to serve as a tool for social science practitioners in analyzing attitudes towards complex social issues \textit{at scale}.

\section*{Acknowledgements}
We would like to thank Julia Mendelsohn, Xuhui Zhou, Rob Voigt, Justin Cho, Marco Castro, Sunyoung Lie, Aida Davani, and Vinodkumar Prabhakaran for their valuable insights on our framing typology and annotation pipeline. 
We would also like to thank the USC NLP faculty, USC Center for AI in Society (CAIS) and students: Sayan Ghosh, Johnny Wei and Matthew Finlayson for their continuous feedback in the formulation of our project and the anonymous reviewers and area chairs in the ARR OpenReview cycle for their helpful suggestions. 
This research was supported by an award from the Allen Institute for AI and a USC Zumberge DEI in Research Fellowship. Brihi was supported by the USC + Amazon Center on Secure \& Trusted Machine Learning. 

\section*{Limitations and Ethical Considerations}

\paragraph{LLMs as Annotation Assistants}
One limitation of using LLMs as annotation assistants for socially sensitive tasks is the potential for harmful biases to be reflected in the predictions. LLMs have been shown to reflect biases \cite{feng2023pretraining, gallegos2023bias} that are considered harmful to underserved communities \cite{kiritchenko-etal-2023-aporophobia}. We are still understanding the capabilities of LLMs in contextualizing and reasoning about sensitive topics such as homelessness that are contingent on complex social and political norms. As a result, we ensure our annotations are validated by experts in social science to mitigate the propagation of harms reflected by LLMs. 
An interesting avenue for future work would be to more effectively scale up large scale annotations for socially sensitive settings by better allocating expert intervention where necessary. 

\paragraph{Diversity of Attitudes}
Given that we do not know the demographic background or geographical origin of the post, our framing typology 
may not necessarily be representative of attitudes towards homelessness outside the U.S. 
Homelessness is viewed differently depending on regionality, ideology, political and economic status among other socio-political factors that might not be represented in our dataset. 
Our framing typology may have introduced our own prior biases in reasoning about these posts that may not be in direct alignment with the general public's attitudes towards homelessness.\footnote{Our annotation team consisted of domain and trained experts from social work and computer science from a mix of backgrounds including 4 from the U.S., 1 from India, and 1 from China.}
The use of LLMs could further exacerbate societal biases in reasoning about an already marginalized population. 
This is an important consideration while using \frames in making generalizations about attitudes towards homelessness. 

\paragraph{Data Collection via Keyword Search}
Our collection of Twitter posts was restricted to those that contained the keyword `homeless' but there are many other terms to reference PEH, which could be explored in future work.\footnote{ 
Initially, we started with a much larger set of seed words including: “homeless, homelessness, unhoused, unsheltered, encampment, tent, houseless, living on the streets, hobo, doss, bag lady, mendicant, couchsurf, housing benefit, street people, soup kitchen, roofless, vagabond, down-and-out, destitute”, 
but we found that this returned content not necessarily related to homelessness.} 

\paragraph{Privacy} Due to the diversity of our dataset and lexical diversity of social media posts, we anonymized user handles, and did not obtain or use any priors about the user from their profile or follower network; however, prior work has shown that demographic background could be useful \cite{tsai2019public}. 
We only rely on the content of the post and the timestamp (time at which the post was created) for our study.

\paragraph{Confounders}
We did not consider additional social, political and economic confounders that can point to why some attitudes are more salient than others. 
For example, substance use and mental illness may influence the public’s opinion about PEH 
\cite{inbook}. 
\frames can serve as the foundation for identifying such factors which can further corroborate the statistical prevalence of certain attitudes over others. Furthermore, taking into consideration the target group of the post may provide more finer grained signals as to who the attitude is directed at and help better classify harmful rhetoric towards PEH \cite{kiritchenko-etal-2023-aporophobia}.  
Initially, we considered the characterization of these additional factors which are shown in Figure \ref{fig:additive}, but it’s extremely difficult and laborious to label posts at scale with more than 50+ frames and labels.  We leave it up to future work to characterize additional social and political factors that can provide important and necessary context in elaborating the statistical prevalence of the frames in our dataset.

\bibliography{custom}

\appendix
\section{\frames Discovery}
\label{sec:appendix:frames}
\begin{figure}[h!]
 \centering
 \includegraphics[width=0.48\textwidth]{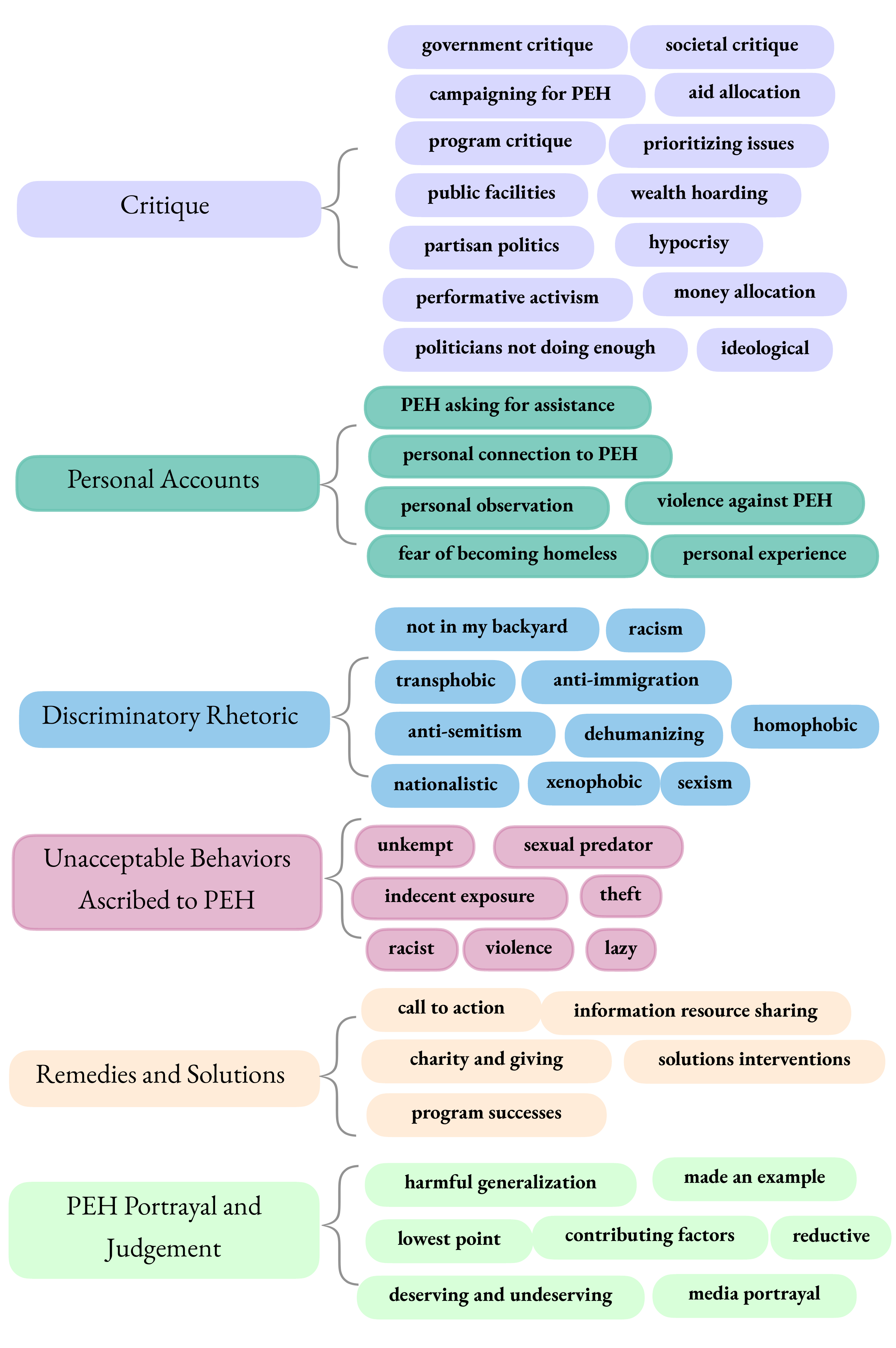}
 \caption{\textbf{\frames v1: } Our first iteration of \frames included more themes and \finer frames. We went applied principles from grounding theory to iterate on our typology to more closely capture public opinions and attitudes towards homelessness.}
 \label{fig:frames_v1}
\end{figure}

\begin{figure}[]
 \centering
 \includegraphics[width=0.42\textwidth]{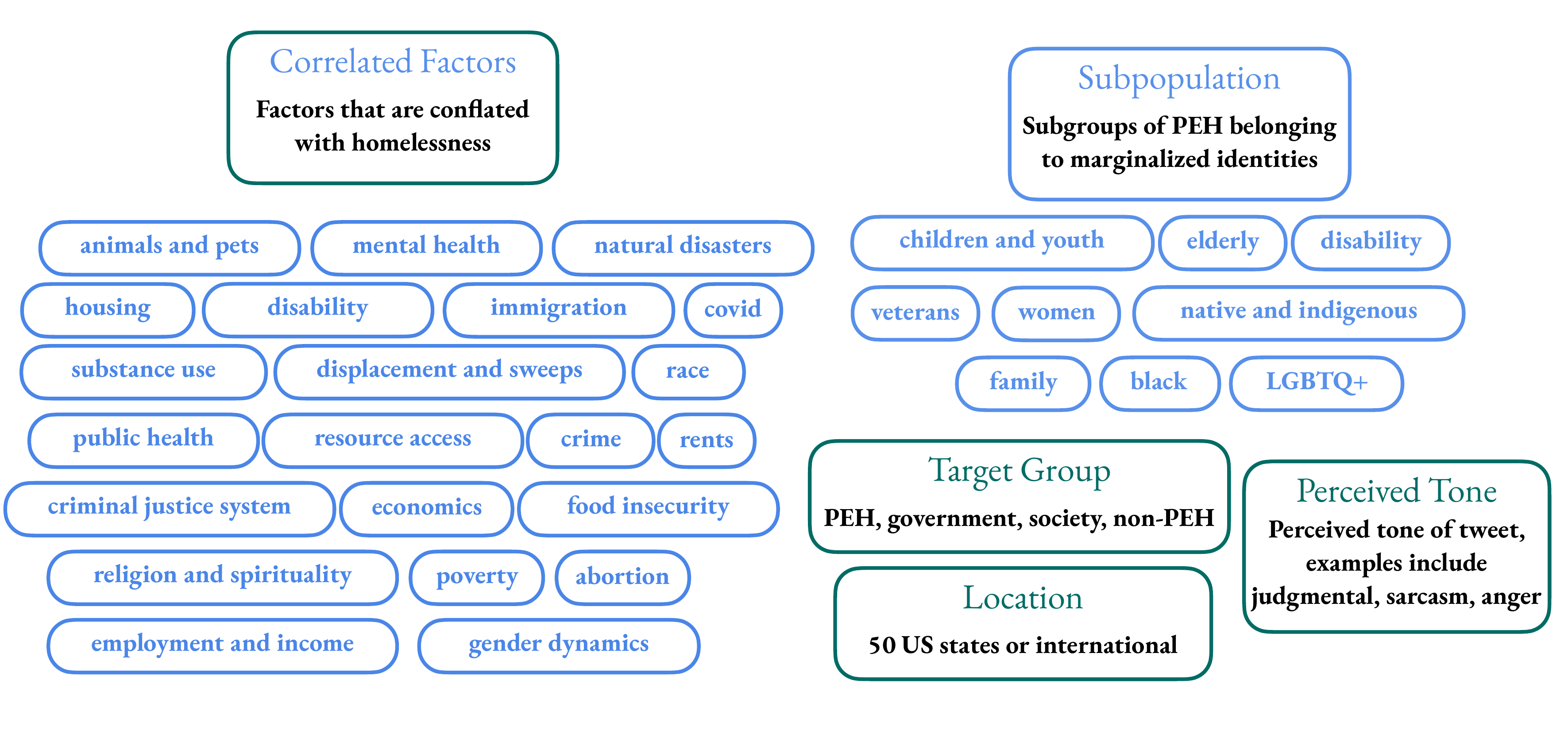}
 \caption{\textbf{Additive Labels:} We also considered additive labels that further contextualized the post with correlated factors often discussed with homelessness. We plan to incorporate additive labels in future work to further understand the social context of attitudes towards homelessness.}
 \label{fig:additive}
\end{figure}

We provide an overview of an earlier iteration of \frames in \autoref{fig:frames_v1} where we had close to 50 \finer frames across six different themes. Following the `coding' process from grounded theory \cite{groundingtheory}, we refined our frames after consensus discussions. 
In our primary iterations, we also considered annotating factors that are often conflated with homelessness. Examples of such factors include substance use and mental illness. Such factors provide important context for how public opinion can be influenced by the social and economic climate of a particular region. We provide details of these factors in Figure \ref{fig:additive} and leave this up to future work as an extension of \frames. In the following section, we outline our  expert annotation \guidelines developed during \textit{Frame Discovery} that contain clarifications of our definitions, common questions that came up during the annotation process and steps on how to address edge cases. We find that incorporating such insights from domain experts in our prompt refinement guided by CoT explanations improves annotation performance on our task. 
\subsection{\guidelines}
\label{sec:appendix:reminders}
\begin{enumerate}
    \item References to homelessness as lowest point, or linguistic structure of “listing”: ‘the country is in shambles, immigration sucks, inflation is up, and homeless crisis is up’ →  are \governmentcritique
    \item Posts that fall under \harmfulgeneralization are references to homelessness as lowest point, usually these posts read `i’d rather be homeless than ...' 
    \item \solutionsinterventions can be expressed in many different ways: volunteering, advocacy etc. If someone is defending PEH from harmful stereotypes, tag with both \solutionsinterventions and \harmfulgeneralization
    \item If someone is advocating against a specific harmful stereotype – label as \harmfulgeneralization and \solutionsinterventions, otherwise, neutral sentiments and positive comments about PEH can be labeled as \personalinteraction
    \item \harmfulgeneralization coupled with \solutionsinterventions means defending against a specific harmful stereotype
    \item \harmfulgeneralization  on its own or coupled with \personalinteraction means person writing the post is making a harmful stereotype 
    \item If no harmful stereotype is mentioned and person is just making a positive comment about PEH: \personalinteraction, possibly \solutionsinterventions
    \item Metaphors/similes ascribing homelessness with negative qualities should be tagged with \harmfulgeneralization
    \item If comparing homelessness to other issues and discussing prioritization of issues: \deservingundeserving, potentially \solutionsinterventions
    \item `looking homeless’ implying PEH have a dirty, unkempt appearance is \harmfulgeneralization
    \item There is a difference between one experience of PEH doing something harmful vs. generalizing this attribute to PEH as a whole. \harmfulgeneralization referring to generalizations made about PEH population as a whole as opposed to one incident. 
    \item  If specifically discussing how money, aid or resources should be allocated across PEH vs. other groups such as immigrants or refugees, tag with \moneyaid
    \item \governmentcritique also include posts which point out political parties by name explicitly → you’re looking for a juxtaposition between parties, for e.g using democratic ideals specifically to critique the party’s response to homelessness in comparison to republican ideals 
    \item \solutionsinterventions includes calls to action which are more directed and pointed and often more declarative in nature, For ex. ‘Hurry up govs!’ or includes critiquing or discussing specific solutions 
    \item Statements like: “the city is so great, why do you have homelessness” should be tagged with \governmentcritique 
    \item When the post is critiquing a solution or is questioning a solution of intervention, it should be augmented with additional labels from the \critiques theme along with \solutionsinterventions. 
    \item If there is critique of solutions funded by the government or government sponsored, then label as \governmentcritique    
    \item Google names of politicians if they show up (\governmentcritique frame often names politicians, if you suspect that’s what they’re doing and don’t recognize the name, then google to confirm) 
    \item Don’t open links to external sources 
    \item If the post mentions policing of homelessness, consider tagging with \governmentcritique since this is government mandated 
    \item Critiquing of how the ultra-rich are donating money is \moneyaid
    \item discussion/debate around contributing factors to homelessness: \solutionsinterventions potentially 
    \item If someone is defending PEH from harmful stereotypes, tag with both \solutionsinterventions and \harmfulgeneralization 
    \item \harmfulgeneralization on it own or coupled with \personalinteraction → person is making a harmful stereotype
    \item If author of the post is expressing prejudice against PEH → tag with \harmfulgeneralization. If the author of the post is calling out prejudice, tag with \harmfulgeneralization and \solutionsinterventions 
\end{enumerate}

\section{Annotation of \frames}
\label{sec:appendix:annotations}
We show our dataset splits for each annotation strategy in \autoref{tab:appendix:annotation_splits}. 

\begin{table}[h!]
\centering
\scalebox{0.7}{
\begin{tabular} {{p{0.4\linewidth} p{0.4\linewidth}p{0.2\linewidth}p{0.2\linewidth}}}
\toprule
\textbf{Annotation Method} & \textbf{Posts} & \textbf{Filtered Out} & \textbf{Remaining} \\
\midrule
\expert & 5.3k & 1.2k& 4.1k\\
\midrule
\gptsolo & 10k & 3k& 7k\\ 
\midrule
\gptintheloop (expert validated) & 4.4k $\in$ 7k \gptsolo & 300 & 4.1k\\
\midrule
\flanlarge (model predictions) & 3.1M & 700k& 2.4M\\
\bottomrule
\end{tabular}
}
\caption{\textbf{Annotation Splits}: We include the number of posts originally annotated, the number of posts filtered out and the number of posts remaining for analysis. 
}
\label{tab:appendix:annotation_splits}
\end{table}

\begin{figure}[h!]
 \centering
 \includegraphics[width=0.48\textwidth]{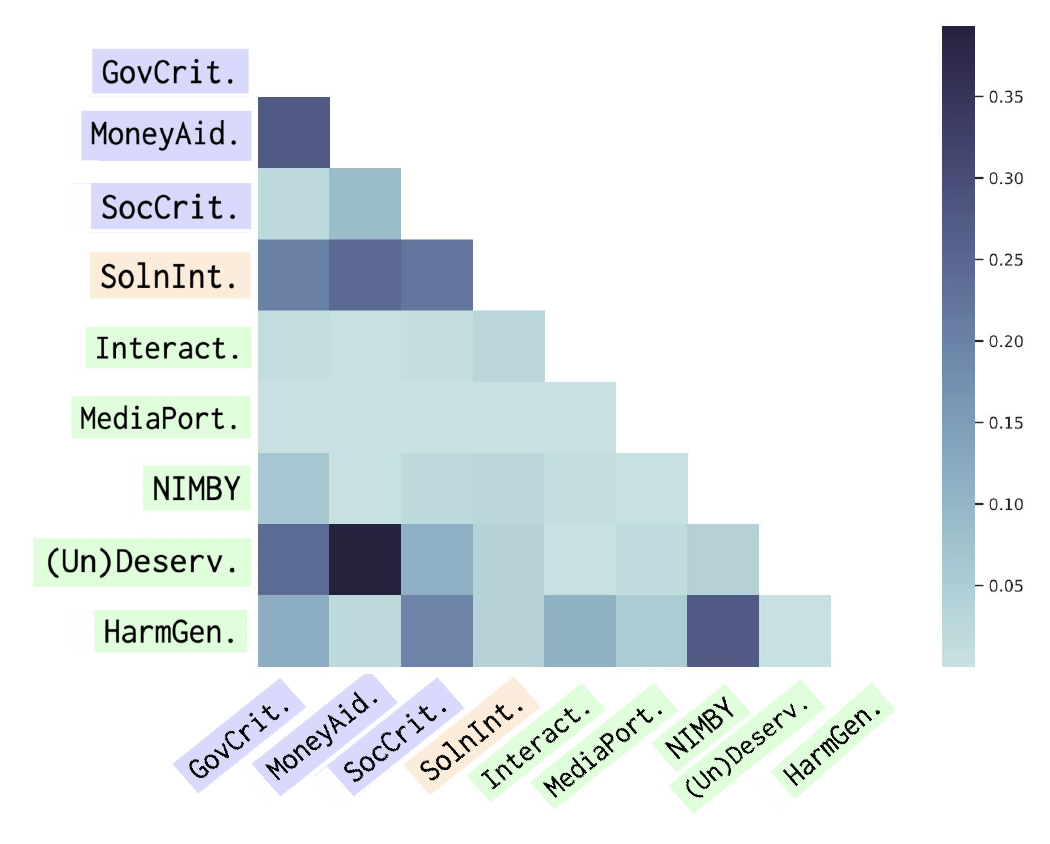}
 \caption{Proportion of posts containing co-occurrences of \finer frames in our set of 2.4M annotated posts. (Each cell is normalized by the column).
 }
 \label{fig:co_occurrence}
\end{figure}

\begin{figure}[h!]
 \includegraphics[width=0.48\textwidth]{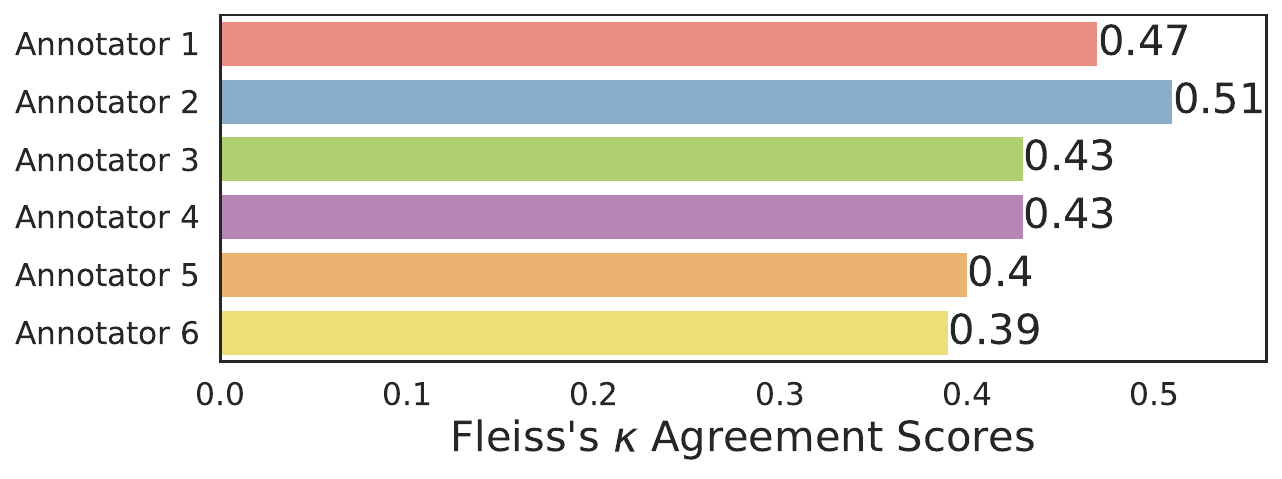}
 \caption{ \textbf{Inter-annotator Agreement Scores}: We plot the average Fleiss' $\kappa$ agreement scores for each expert (Annotators 1\ldots6) where each Annotator's Fleiss' $\kappa$ is computed by using every other annotator as a reference.
 }
 \label{fig:agreement_scores}
\end{figure}

\begin{table*}[h!]
\centering
\resizebox{\linewidth}{!}{
\begin{tabular}{p{0.99\linewidth} p{0.4\linewidth}}
    \toprule
    \textbf{Post} & \textbf{Expert Annotations}\\ 
    \midrule
    {\fontfamily{cmss}\selectfont we used to have a fella come to our late night hockey games, sit in the stands and knit, one night we asked what he was knitting any why he chose a hockey arena, his response was he enjoyed watching the ladies play, and knitting scarfs and hats for the homeless was his goal.} & \solutionsinterventions \\
     \midrule
     {\fontfamily{cmss}\selectfont yes, there are rich people who own companies like tesla, who employ thousands of people who are a plus for our economy. they work, they get paid, they buy houses, cars, groceries, become citizens of law abiding communities, etc. they feel good about themselves. no jobs, homeless.} & \solutionsinterventions, \moneyaid \\
     \midrule
     {\fontfamily{cmss}\selectfont borders were legally open (only need single lane), business closed cuz they were told to (one place tried to stay open \&amp; got harassed by police, homeless were fed \&amp; the truckers were the ones harassed, egged \&amp; called names.} & \deservingundeserving \\
     \midrule
     {\fontfamily{cmss}\selectfont considerable damage has already been done this isnt just for pain medication too bc so many black, brown, trans,, queer poor, disabled (lol), homeless, immigrant, addicts, mentally ill, and beyond (most people in general) domt get believed by doctors and get pawned off} & \harmfulgeneralization, \societalcritique \\
     \midrule
     {\fontfamily{cmss}\selectfont what about our rights to a secure border!?
 those laws were passed by a democratic process. yet you violate your oath of office and give aid and comfort to those that break our laws.
 use funds to aid them,that could give our citizens m4all!
 look at the homeless in your state!!!} & \governmentcritique, \deservingundeserving, \solutionsinterventions, \moneyaid \\
     \midrule
    {\fontfamily{cmss}\selectfont now, shall we show the millions of refugees russia created? tens of millions of homeless? yeah. russia staged a photo op where everything is fine while mariupol of half a mil has been destroyed completely by russia. good job on the ethical balance. murderers can be friendly.} & \governmentcritique \\
     \midrule
    {\fontfamily{cmss}\selectfont your not going to do anything, what makes you think most the homeless want to work, pay rent, pay taxs, most can't maintain long enough to hold or get a job. can't do drugs on the job. they would rather sit in a tent getting high. she loves her meth to much.} & \harmfulgeneralization \\
     \midrule
     {\fontfamily{cmss}\selectfont i have a couple of ideas for the elderly homeless, and the homeless vets.
 too radical for most people, but, you might be interested to hear them.} & \solutionsinterventions \\
     \midrule
     {\fontfamily{cmss}\selectfont im not saying he is guilt or anything, i hanged around a lot ot of homeless people when i was still in high school, they are an odd bunch of people, they had public next to their sleeping, however no one used it everyone was shiting everywhere but inside the toilet} & \harmfulgeneralization, \personalinteraction \\
     \midrule
     {\fontfamily{cmss}\selectfont covid; mislead you, talk down the seriousness of the pandemic or letting it spread, gop, top 1\%, and their lobbyist theory was; less people more money can go around, believing it would kill most of the homeless or unhealthy people that are a burden to the healthcare sector.} & \harmfulgeneralization, \moneyaid, \governmentcritique, \deservingundeserving, \solutionsinterventions \\
     \midrule
     {\fontfamily{cmss}\selectfont my impression is that homelessness is highly comorbid with psychiatric illness, so i would *ideally* like to see the homeless being fed as if they were being treated for metabolic syndrome. (i'm actually involved in such an initiative, but we're temporarily on hold.)} & \harmfulgeneralization,\personalinteraction \solutionsinterventions \\
     \midrule
     {\fontfamily{cmss}\selectfont that guy his name is joey. he was part of an organization called center for political innovation, headed by a clown named caleb maupin but apparently lost all of his money, and became homeless, but still defense caleb till the end.} & \personalinteraction, \governmentcritique \\
     \midrule
     {\fontfamily{cmss}\selectfont homeless to the shelters, edp's to the mental hospitals.} & \solutionsinterventions \\
     \midrule
     {\fontfamily{cmss}\selectfont why in the world are you giving homeless folks \$12,000 a piece? do denver voters know you are doing this? you are asking for drug addicts and drunks on the streets making the city look like a third-world country. another disgraceful democrat mayor.} & \harmfulgeneralization, \moneyaid, \governmentcritique, \nimby \\
     \midrule
     {\fontfamily{cmss}\selectfont man, that's some epic rambling bullsh** right there. what is it that you idolize about that guy? i mean, seriously, you could get the same level of coherence out of a random meth-head from the homeless shelter.} & \harmfulgeneralization, \societalcritique \\
     \midrule
     {\fontfamily{cmss}\selectfont here we are - back to people proving who they are. should homeless people get a chance to vote? people who ran away from home and don’t have a birth certificate? why can you buy a gun w/o an id, but you need one to vote? it’s because the gop wants guns, but don’t want poc voting.} & \harmfulgeneralization, \governmentcritique, \solutionsinterventions \\
     \midrule
     {\fontfamily{cmss}\selectfont seriously?? he’s saying the city is spending \$170 million on bike lanes, while the homeless count soars.
 \$170m is better spent pretty much on anything else.} & \moneyaid \\
     \midrule
     {\fontfamily{cmss}\selectfont and the people who still have to work, even if they get heatstroke? the homeless? those in poor housing? not everyone has the freedom to defend themselves against extreme heat - or a sociopath government and public discourse.} & \societalcritique, \governmentcritique, \solutionsinterventions \\
     \midrule
     {\fontfamily{cmss}\selectfont i literally can't stand when ppl record themselves doing something for someone in need, like giving food to homeless ppl. it isn't ur situation to profit off of my luv} & \societalcritique \\
    \bottomrule 
\end{tabular}
}
\caption{Examples of expert annotated posts with \frames.}
\label{tab:appendix:appendix_examples}
\end{table*}

\begin{figure*}[h!]
\centering
 \includegraphics[width=0.9\textwidth]{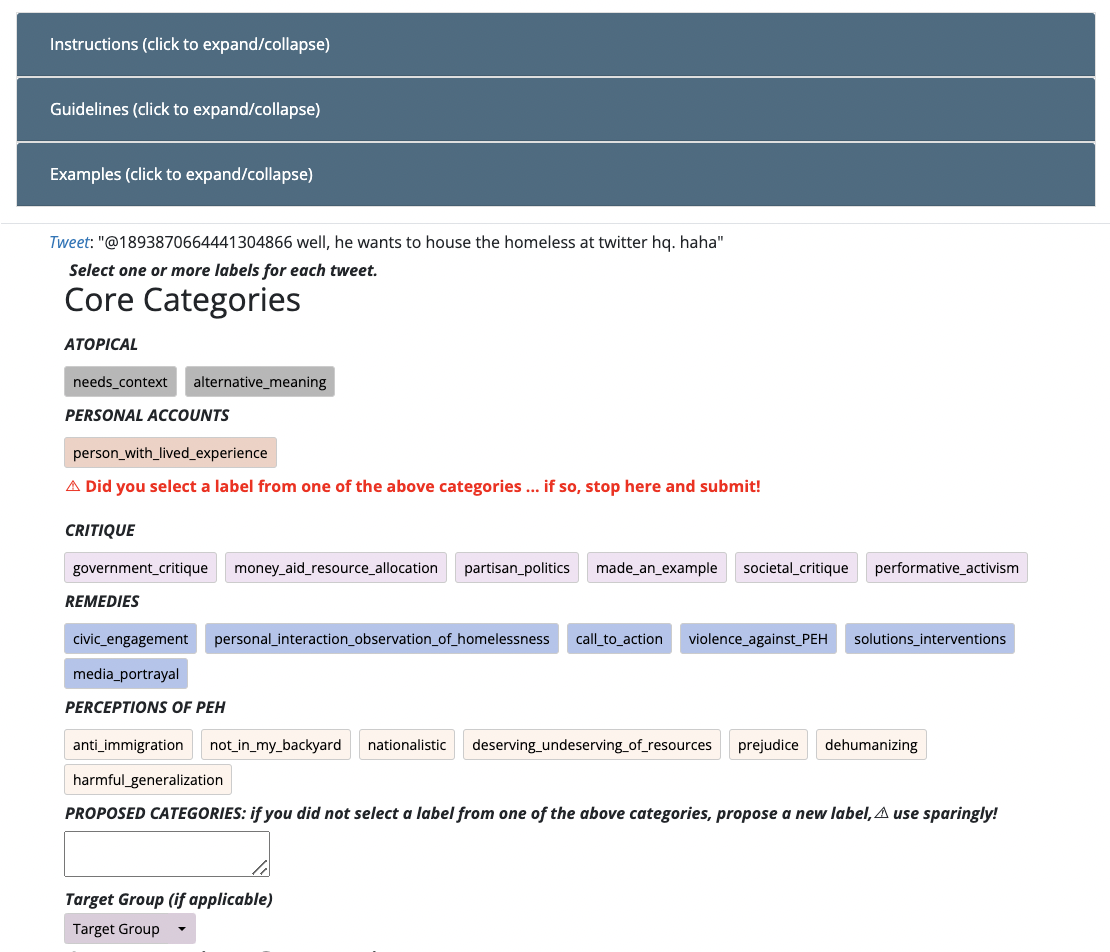}
 \caption{\textbf{Amazon Mechanical Turk Sandbox Interface for Domain / Trained Expert Annotation.} Each annotator is asked to select labels that apply the most to a given post. The time taken for each annotation is also recorded in the backend. This interface details an earlier iteration of \frames that included a larger number of \finer frames.}
 \label{fig:turk}
\end{figure*}

\subsection{Expert Annotations}
\label{sec:appendix:human_annotations}
54\% of our posts are labeled with more than one \finer frames. We detail the co-occurrence of our frames in \autoref{fig:co_occurrence} in our set of 4.1k expert annotated posts. We find that \deservingundeserving and \moneyaid frequently co-occur.  The time period of our posts is from 2021-2023, and in this time, a main agenda for the republic party was sending asylum seekers across the country, predominantly to blue states; this resulted in a lot of immigrants becoming homeless and also resulted in a news cycle that pitted immigrants against PEH in competition for jobs, housing and funding \cite{Abdalla_2022}. Furthermore, \governmentcritique and \moneyaid frequently co-occur as critique against government institutions and politicians is often in the context of funding and resource allocation. \harmfulgeneralization co-occurs frequently with \nimby and \personalinteraction. \nimby rhetoric can be expressed in harmful and exclusionary ways towards PEH, and furthermore, personal interactions with PEH can elicit more harmful stereotyping.

We provide details of our inter-annotator agreement scores in \autoref{fig:agreement_scores} and an example of our Amazon Mechanical Turk Sandbox annotation interface in \autoref{fig:turk}. The expert first determines whether the post should be filtered out if it is not relevant to public attitudes towards homelessness. For relevant posts, the expert must select at least one or more \finer frames to label the post. We provide examples of our expert annotated posts in Table \ref{tab:appendix:appendix_examples}.

\subsection{LLM-Assisted Expert Annotation}
\label{sec:appendix:gpt_annotations}

We include the final prompt that we achieve after iterating through the \frames instructions, for obtaining \gptsolo annotations (\autoref{tab:gpt_prompt}).

Our prompt consists of a two-step process: (1) Using the \filter prompt to filter relevant posts, followed by (2) Using \frames to get multi-label \finer frames for each post. 
We provide examples of using CoT and expert observations to incorporate \guidelines in \autoref{tab:gpt_cot}. 
We also included examples of frames with the highest false discovery rates and highest precision as predicted by \gptsolo in \autoref{tab:gpt4_error_analysis}.

\begin{table*}[h!]
\centering
\scalebox{0.6}{
\begin{tabular}{p{23cm}}
\toprule
\midrule
\textbf{\filter prompt}\\
\textbf{System Prompt}: 
You are an AI model trained to classify tweets related to homelessness into 2 different labels. The labels are $\langle$attitude\_towards\_homelessness$\rangle$ and $\langle$other$\rangle$.

Label Descriptions:
- $\langle$attitude\_towards\_homelessness$\rangle$:  Includes tweets about homelessness to talk about another topic or are generally about the social issue of homelessness. They cover a range of topics related to homelessness, including criticism of government bodies, institutions, or political parties, discussions about the allocation and disbursement of money, aid, and resources, criticism of societal attitudes towards homelessness, debates about who is more deserving of resources, harmful generalizations or stereotypes about homeless people, opposition to having homeless people in local areas or neighborhoods, references to media portrayals of homelessness, anecdotes about interactions with homeless people, and suggestions or ideas for solving the homelessness crisis.

- $\langle$other$\rangle$: Includes personal anecdotes from people experiencing homelessness who are sharing their personal experience while being homeless or asking for assistance and aid. This category does NOT include tweets about fictional characters and personal interactions with other homeless people. Includes statements that are nonsensical or difficult to decipher and require access to additional resources like links, media, images, etc in order to properly interpret the tweet or references to homeless animals or being politically homeless.\\
\textbf{Prompt}: Classify the following tweet into one of the provided labels:

"post"

In concise points, please provide the relevant label that best characterizes the content of the tweet. Do not "read into" the text with interpretations, stick to the definitions of the categories strictly. The format should be the predicted label, followed by "because", followed by reason. Do not add any additional text.

Feel free to reference the label descriptions to support your classification. Provide any relevant context that influenced your classification. \\
\midrule
\textbf{\frames prompt}\\
\textbf{System Prompt}: 
You are an AI model trained to classify tweets related to homelessness into 9 different labels. The labels include $\langle$government\_critique$\rangle$, $\langle$money\_aid\_resource$\rangle$, $\langle$public\_critique$\rangle$, $\langle$deserving\_undeserving\_of\_resources$\rangle$, $\langle$harmful\_statements\_against\_homelessness$\rangle$, $\langle$not\_in\_my\_backyard$\rangle$, $\langle$media\_portrayal$\rangle$, $\langle$interaction\_with\_homeless\_person$\rangle$, and $\langle$solutions\_interventions$\rangle$.

Label Descriptions:

- $\langle$government\_critique$\rangle$: criticism about the government body, government institutions or political parties including critique of specific politicians, policies about homelessness, critique of programs that are being funded or considered by the government such as welfare programs, and the policing of homelessness. Also includes statements where homelessness is used as a vehicle or stand-in to talk about a broader issue portraying homelessness amongst other negative social and government problems in a list-like manner in a tweet like "murder rates, homelessness, immigration and inflation. all suck". Also includes statements that mention names of politicians.

- $\langle$money\_aid\_resource$\rangle$: Primarily includes discussion of money, for long term relief of homelessness. Includes aid or resource disbursement and allocation by government, institutions, organizations or wealthy individuals (not regular public) and also includes discussion or critique and suggestions on how the government decides to spend money and resources. Also includes discussions of giving or providing money, aid and resources to homeless people.

- $\langle$public\_critique$\rangle$: Criticism of society in general or social norms that includes discussion of society at large instead of specific people, often pointing out hypocrisy and critiquing society’s general attitudes towards homelessness. Also includes critiquing someone helping homelessness in order to gain some personal benefit where someone is being explicitly called out for doing charitable acts while filming a video or for recognition.

- $\langle$deserving\_undeserving\_of\_resources$\rangle$: Discussion of competing priorities where homelessness is compared to other issues that more or less deserve aid and resources. Includes statements that express anti-immigration and support for policies, political initiatives and actions that restrict immigration often comparing and prioritizing aid to people experiencing homelessness over immigrants. Also includes nationalistic statements that prioritize one’s own nation over others including discussion about prioritizing aid and relief for veterans and the nation’s citizens over non-citizens. 

- $\langle$harmful\_statements\_against\_homelessness$\rangle$: Blanket statements that generalize a negative, harmful or undesirable attribute to all people experiencing homelessness and invoke stereotypes and make assumptions about people experiencing homelessness as a whole. Examples include statements that say all people experiencing homelessness are violent, addicts, thieves, mentally ill, unkempt, dirty, and poor at managing finances and also comparing dirty, disheveled clothing to ‘looking homeless’. Includes statements that express prejudice against homelessness such as sexism, homophobia, racism, anti-semitism and transphobia or dehumanize people experiencing homelessness depriving them of positive human qualities and viewing them as sub-human or as trash. Includes statements that portray homelessness as the lowest point in one’s life where homelessness is used as an example of something wrong or bad. This also includes metaphors to describe objects like anti-homeless. Could also include statements that express the desire to be violent strictly against people experiencing homelessness including threats against homelessness. Includes statements that portray homelessness as the lowest point in one’s life where homelessness is used as an example of something wrong or bad. This includes listing homelessness in conjunction with other issues that are viewed as problematic or negative.

- $\langle$not\_in\_my\_backyard$\rangle$: Opposition by residents to proposed developments in their local area, as well as support for strict land use regulations against wanting to see homelessness in their local area and neighborhood. Also includes displacement sweeps to remove PEH from certain areas and neighborhoods. 

- $\langle$media\_portrayal$\rangle$: Reference to a fictional character that is portraying homelessness and includes tweets and links about local news media. 

- $\langle$interaction\_with\_homeless\_person$\rangle$: Only includes anecdotes describing a real-life interaction with a homeless person.

- $\langle$solutions\_interventions$\rangle$: Suggestions, remedies, problem solving and ideas for alleviating the homelessness crisis including support for policy reform, existing policies and welfare programs. Includes individual people giving money, food and help for immediate relief of homelessness. Also includes charitable acts, non-profit work, providing help and emergency aid relief, and defending people experiencing homelessness from harmful stereotypes and generalizations and advocating for positive qualities for people experiencing homelessness. Also includes call to action statements that invoke a sense of urgency in taking action towards helping the homelessness crisis. \\
\textbf{Prompt}: Classify the following tweet into one or more of the provided labels:

"post"

In concise points, carefully assess the relevant label(s) that best characterize the content of the tweet; try to list all the labels that are applicable for the tweet. Do not "read into" the text with interpretations or indications or make any assumptions, and stick to the definitions of the labels strictly. Each individual label should be followed by "because", followed by the reason for why that label was picked. Do not add any additional text. You have to select atleast one label, you cannot leave it out.

Feel free to reference the label descriptions to support your classification. Provide any relevant context that influenced your classification. \\

\bottomrule
\end{tabular}
}
\caption{\textbf{\gptsolo prompts}: We use the above \filter and \frames prompt in sequence, to obtain \gptsolo labels. 
}
\label{tab:gpt_prompt}
\end{table*}
\begin{table*}[h!]
\centering
\resizebox{\linewidth}{!}{%
\begin{tabular}{p{4cm}p{20cm}}
\toprule
\textbf{\finer Frames}&\textbf{Examples}\\
\midrule
\makecell[l]{\textbf{\deservingundeserving}\\\textbf{(Highest False Discovery)}} &
\makecell[l]{
\textbf{Post}: \@mention we have 600,000 homeless and you sound self-centered. typical capitalist mindset.
\\
\textbf{Predicted CoT}: Because the post mentions the number of homeless people, implying that resources should be\\ allocated to them, and criticizes the individual for not prioritizing this issue.\\
\textbf{Observation}: The post uses homelessness as an example to critique an individual, \\ but the CoT makes implications of PEH being in competition for resources when \\there is not enough information in the post to make that claim. \\
\midrule
\textbf{Post}: \@mention furthermore this city has at least three homeless shelters available for these people\\ - they just have to stay sober.
\\
\textbf{Predicted CoT}: Because the post implies that homeless people are undeserving of shelter if they can't stay sober,\\ suggesting a condition for them to receive resources.\\
\textbf{Observation}: The post does not imply that PEH are in competition for resources, rather it is making a harmful\\ generalization about substance use.  
}\\

\midrule
\makecell[l]{\textbf{\personalinteraction}\\\textbf{(Highest False Discovery)}}&
\makecell[l]{
\textbf{Post}: \@mention are you helping homeless british people including veterans?
\\
\textbf{Predicted CoT}: Because the post is addressing someone (presumably a person or organization) about their actions\\ towards homeless people, which can be seen as a form of interaction.\\
\textbf{Observation}: It is not clear if this post is talking about a one on one interaction with PEH, the CoT is generalizing\\ and reading into the post. \\
\midrule
\textbf{Post}: \@mention wonder if any one here has\\ offered up a room to the many thousands of irish homeless
\\
\textbf{Predicted CoT}: Because the post is suggesting that the individuals addressed could have a direct interaction with\\ homeless people by offering them a room\\
\textbf{Observation}: The post does not mention a direct interaction with PEH and the CoT implies that \gptsolo is reading\\ into the post. 
}\\

\midrule
\makecell[l]{\textbf{\harmfulgeneralization}\\\textbf{(Highest Precision)}}&
\makecell[l]{
\textbf{Post}: \@mention how are they a nuisance? i'd rather have that than the homeless;\\ their garbage littering public spaces.
\\
\textbf{Predicted CoT}: Because the post generalizes all homeless people as a nuisance and associates them with littering\\ public spaces, which is a negative stereotype.\\
\textbf{Observation}: The CoT correctly identifies that PEH are all being portrayed as a nuisance which can be considered\\ a negative stereotype. \\
\midrule
\textbf{Post}: \@mention furthermore this city has at least three homeless shelters available for these people\\ - they just have to stay sober. they’re out there by their f*cken choice
\\
\textbf{Predicted CoT}: Because the post generalizes that all homeless people are choosing to stay on the streets due to\\ their inability to stay sober, which is a harmful stereotype.\\
\textbf{Observation}: The CoT reasoning correctly identifies the harmful generalization about PEH and substance use, and\\ that all PEH are responsible for their own plight. \\
}\\

\midrule
\makecell[l]{\textbf{\governmentcritique}\\\textbf{(Highest Precision)}}&
\makecell[l]{
\textbf{Post}: \@mention \@mention \@mention \@mention ur\\ obviously not from the city cause bratton is a notorious bigot who hates the homeless, blacks and the poor. he said\\ what he said and he meant what he meant... and what he meant was “homeless people are a nuisance.” his record\\ speaks for itself. but go on...
\\
\textbf{Predicted CoT}: because the post criticizes a specific government figure, Bratton, for his alleged negative attitudes\\ towards the homeless, blacks, and the poor. The user is critiquing his policies and statements about homelessness. \\
\textbf{Observation}: The CoT reasoning correctly identifies critique against William Bratton, the NYC law enforcement\\ officer. \\
\midrule
\textbf{Post}: \@mention \@mention big ron is the one who destroyed the institutions that\\ held our mentally ill. he put them on the streets that added thousands of homeless and mentally ill people to live like\\ animals. and people think the republicans are there to help all the conservatives, wrong! just the rich.
\\
\textbf{Predicted CoT}: because the post criticizes a specific politician ("big ron") for his policies that allegedly\\ led to an increase in homelessness and mental illness on the streets. The post also criticizes the Republican party,\\ implying they only help the rich.\\
\textbf{Observation}: The CoT reasoning correctly identifies critique against a politician for his policies and political party\\ values. \\
}\\
\bottomrule
\end{tabular}
}
\caption{\textbf{\gptsolo Error Analysis}: We compare instances that are annotated using \gptsolo and provide the generated Chain-of-Thought explanations, along with expert observations of how \gptsolo is getting these examples incorrect/correct.}
\label{tab:gpt4_error_analysis}
\end{table*}

\begin{table*}[h!]
\centering
\resizebox{\linewidth}{!}{
\begin{tabular}{p{0.3\linewidth} p{0.5\linewidth}p{0.99\linewidth}}
    \toprule
    \textbf{Post} & \textbf{\gptsolo CoT and Expert Observation}& \textbf{Prompt Editing: Frame Definition Clarification}\\ 
    \midrule
    {\fontfamily{cmss}\selectfont what a shi*show, can't wait til your all gone, go volunteer at homeless shelters do some good} & \textbf{\gptsolo CoT:} \societalcritique: because the tweet criticizes a group of people and suggests they should do something beneficial like volunteering at homeless shelters. \newline \textbf{Expert Observation:} The generated CoT is not incorrect, however, \gptsolo was not able to capture the call-to-action in the last part of the post. According to \frames-Guidelines \#14: \solutionsinterventions includes calls to action which are more directed and pointed and often more declarative in nature. We use this guideline to expand the definition of \solutionsinterventions. & \textbf{Original Definition + Addition:} \solutionsinterventions: Suggestions, remedies, problem solving and ideas for alleviating the homelessness crisis including support for policy reform, existing policies and welfare programs. Includes defending people experiencing homelessness from harmful stereotypes and generalizations and advocating for positive qualities for people experiencing homelessness. \textcolor{teal}{Also includes individual people giving money, food and help for immediate relief of homelessness, charitable acts, non-profit work, providing help and emergency aid relief, call to action statements that invoke a sense of urgency in taking action towards helping the homelessness crisis} \newline \textbf{Updated CoT:}\societalcritique because the tweet criticizes unspecified individuals or groups, suggesting they should be doing more to help the homeless, such as volunteering at shelters. \solutionsinterventions because the tweet suggests volunteering at homeless shelters as a way to help alleviate the homelessness crisis. \\
     \midrule
    {\fontfamily{cmss}\selectfont ... hey @mention would it be able to get a better @mention package? you got it set to homeless edition and it's terrible to work with....how am i supposed to use all my crayons?} & \textbf{\gptsolo CoT:} \governmentcritique because the tweet is addressing a specific entity (presumably a government body or institution, given the context) with a critique about a package that is described as homeless edition, implying dissatisfaction with a program or policy related to homelessness. \newline \textbf{Expert Observation:}  It is unclear whether the post is addressing the government; thus we cannot reliably assume it is \governmentcritique. According to \frames-Guidelines \#8: metaphors/similes ascribing homelessness with negative qualities should be labeled with \harmfulgeneralization as being in done in this post with ‘homeless edition’.  & \textbf{Original Definition + Addition:} \harmfulgeneralization: Blanket statements that generalize a negative, harmful or undesirable attribute to all people experiencing homelessness and invoke stereotypes and make assumptions about people experiencing homelessness as a whole. Examples include statements that say all people experiencing homelessness are violent, addicts, thieves, mentally ill, unkempt, dirty, and poor at managing finances and also comparing dirty, disheveled clothing to ‘looking homeless’. Includes statements that express prejudice against homelessness such as sexism, homophobia, racism, anti-semitism and transphobia or dehumanize people experiencing homelessness depriving them of positive human qualities and viewing them as sub-human or as trash. \textcolor{teal}{Includes statements that portray homelessness as the lowest point in one’s life where homelessness is used as an example of something wrong or bad. This also includes metaphors to describe objects like anti-homeless. Could also include statements that express the desire to be violent strictly against people experiencing homelessness including threats against homelessness.} \newline \textbf{Updated CoT:}\harmfulgeneralization because the tweet uses the term "homeless edition" in a negative context, implying that something associated with homelessness is "terrible to work with". This could be seen as a generalization that negatively portrays homelessness. \\
    \midrule
    {\fontfamily{cmss}\selectfont Tragedy after tragedy, we come into work and go throughout our day as conditions worsen before our eyes. Kids are murdered. Physical/mental health issues go untreated. People are unable to afford food \&; housing and many are dying in the streets homeless, sick and in debt.} & \textbf{\gptsolo CoT:} \societalcritique: because the tweet criticizes society's indifference towards worsening conditions, including homelessness and poverty. \newline \textbf{Expert Observation:} All the problems being listed are social problems that can be considered the faults of the government. According to \frames-Guidelines \#1: References to homelessness as lowest point, or linguistic structure of “listing”: `the country is in shambles, immigration sucks, inflation is up, and homeless crisis is up'  are \governmentcritique. We modify the definition using this guideline. & \textbf{Original Definition + Addition:} \governmentcritique: Criticism about the government body, government institutions or political parties including critique of specific politicians, policies about homelessness, critique of programs that are being funded or considered by the government such as welfare programs, and the policing of homelessness. \textcolor{teal}{Also includes statements where homelessness is used as a vehicle or stand-in to talk about a broader issue portraying homelessness amongst other negative social and government problems in a list-like manner in a tweet like "murder rates, homelessness, immigration and inflation. everything is terrible!".} \newline \textbf{Updated CoT:} \governmentcritique because the tweet implies a critique of the government's inability to address worsening social conditions, including homelessness.\\
    \bottomrule 
\end{tabular}
}
\caption{Examples of prompt refinement using insights from domain experts, CoT and \guidelines.}
\label{tab:gpt_cot}
\end{table*}

\section{LLM-Predicted \frames: Training Details}
\label{sec:appendix:prediction}

Combining expert annotated posts along with posts that are annotated by \gptsolo and expert validated using \gptintheloop in \textsection \ref{sec:llm_collaboration}, we obtain a total of 10400 posts.
We then split these posts into a train-validation-test split using the following criteria:
\begin{itemize}
    \item We ensure roughly an 80:10:10 proportion for the above mentioned split.
    \item The test set contains \textit{only} domain-expert annotated posts. We pick 280 posts that were used in our agreement set in \S\ref{sec:oath-frames}, and add another 1000 instances to the pool. This overall set of 1280 posts is used to pick the best performing \flan variation for final inference on a larger set of posts.
    \item The remaining set of domain-expert annotated and \gptintheloop annotated posts are then combined and split into a train and validation split.
\end{itemize}
We use the pre-trained \flan models for the \texttt{conditional generation} task, pooled from the \texttt{HuggingFace\footnote{\url{https://huggingface.co/docs/transformers/model_doc/flan-t5}}} repository.
The task is structured as follows:
\begin{itemize}
    \item \textbf{Input format}: The input format consists of the following instruction, along with the post to be labeled -- \texttt{Classify the given post into one or more of the following 10 labels: \{list of labels\} post: \{\}}
    \item \textbf{Output format}: If a post needs to be filtered out using the \filter filter, we generate a `0' label, otherwise, we generate a comma-separated list of \finer frames. Note, this is slightly different from how \gptsolo predictions are generated where \gptsolo is prompted in a sequential manner to first filter out the posts according to the \filter filter and then predict the \finer frames on the remaining posts whereas we use \flan to predict the\filter filter in conjunction with the 9 \finer frames all at once. 
\end{itemize}

We demonstrate the training details and results of different \flan variations in \autoref{tab:appendix:model_details} and \autoref{tab:res:finer_models_vs_everyone}.

\begin{table}[h!]
\centering
\scalebox{0.75}{
\begin{tabular}{cc}
\toprule
\textbf{Config}&\textbf{Assignment}\\
\midrule
models &
\makecell[c]{
\textbf{\flanxl}\\
Number of parameters: 3B
\\
\midrule
\textbf{\flanlarge}\\
Number of parameters: 770M
\\
\midrule
\textbf{\flanbase}\\
Number of parameters: 250M
\\
\midrule
\textbf{\flansmall}\\
Number of parameters: 80M\\
\midrule
}\\
train batch size&32\\
eval batch size&16\\
seed&0\\
max epochs&25\\
learning rate&3e-5\\
learning scheduler&fixed\\
GPU&Quadro RTX 8000\\
\# of GPUs &8 (training), 20 (inference)\\
Training time& 8 hours\\
Stopping Criteria& Early Stopping on Validation Loss\\
\bottomrule
\end{tabular}
}
\caption{\textbf{Model Training Details:} We provide details about our models that were finetuned on our Domain + Expert data and \gptintheloop data where \flanlarge achieved the highest Macro F$_1$ across all 9 \finer frames. }
\label{tab:appendix:model_details}
\end{table}

\section{Analyzing Posts with \frames}
\label{sec-analysis}

\begin{figure}[h!]
\centering
 \includegraphics[width=0.4\textwidth]{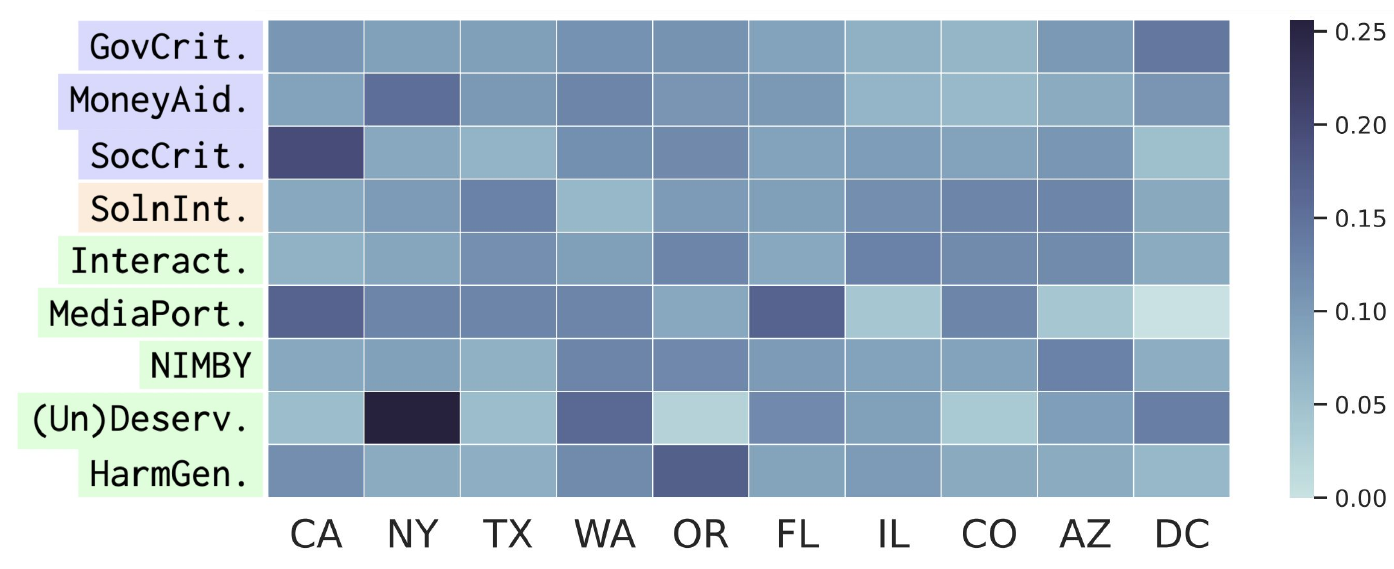}
 \caption{\textbf{Attitudes across state mentions:} 
 Predominant attitudes in posts containing mentions of states labeled with \finer frames as a fraction of total posts labeled with a specific frame (normalizing each cell by the row).
 }
 \label{fig:all_states}
\end{figure}

\begin{figure}[h!]
\centering
 \includegraphics[width=0.4\textwidth]{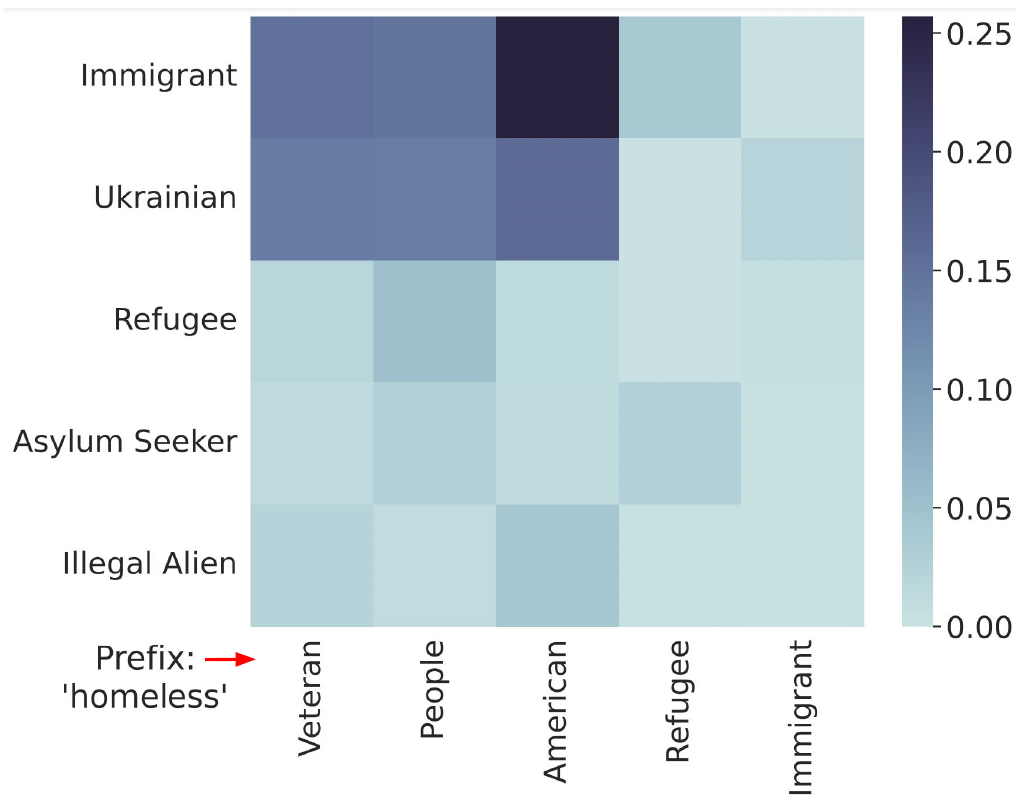}
 \caption{\textbf{Co-occurrence of PEH and vulnerable populations:} Number of posts containing mentions of PEH: `homeless'-(`veteran', `people', `american', `refugee', `immigrant') and vulnerable populations (`Immigrant', `Ukrainian', `Refugee', `Asylum Seeker', `Illegal Alien') as a fraction of the total number of posts mentioning PEH (normalizing each cell by the column).}
 \label{fig:co_occur_identity}
\end{figure}

\begin{figure*}[h!]
\centering
 \includegraphics[width=0.9\textwidth]{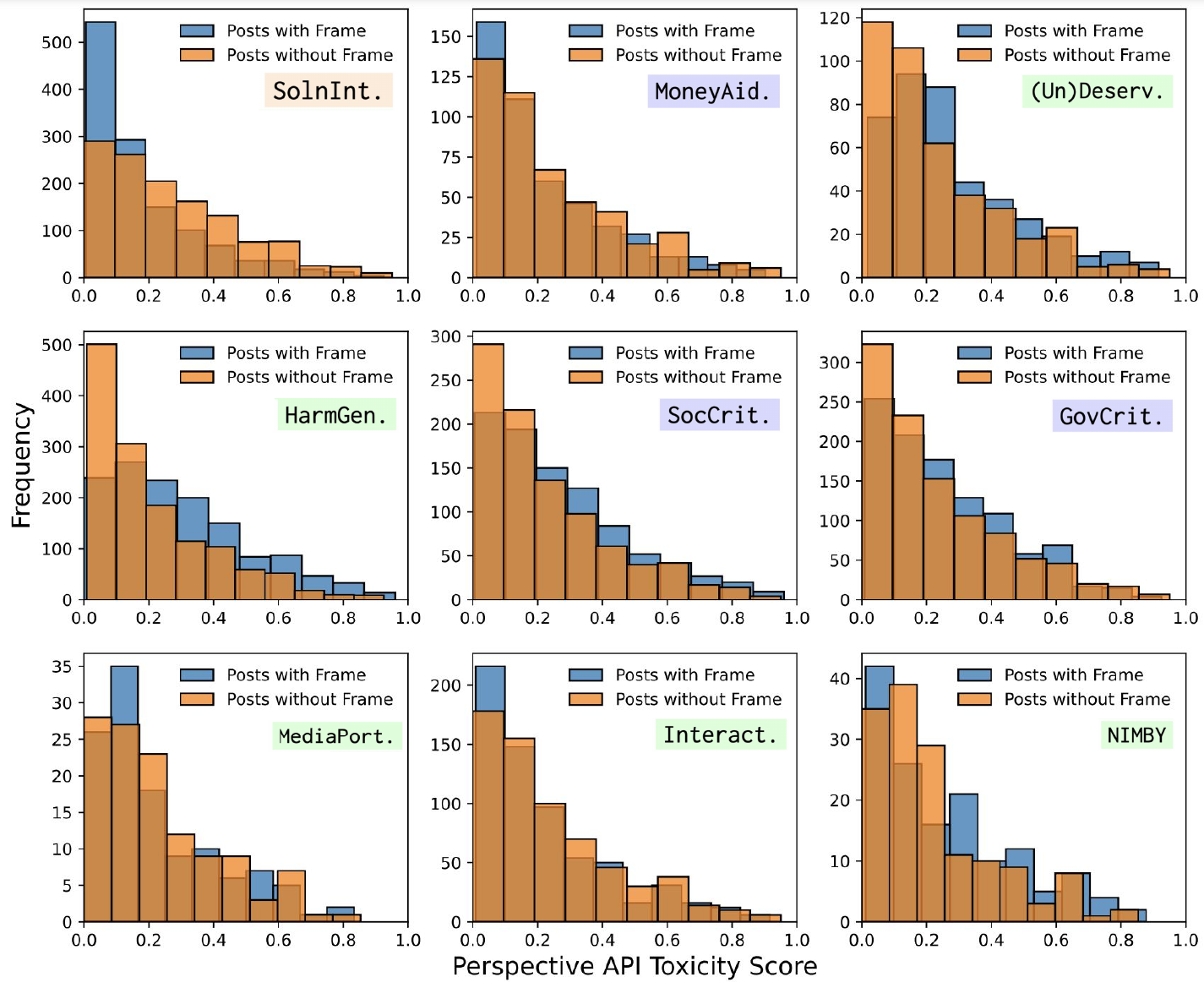}
 \caption{Distribution of toxicity scores from Perspective API toxicity classifier for posts labeled with each \finer frame in our subset of 4.1k human-annotated posts.}
 \label{fig:tox}
\end{figure*}

\begin{figure*}[h!]
\centering
 \includegraphics[width=0.9\textwidth]{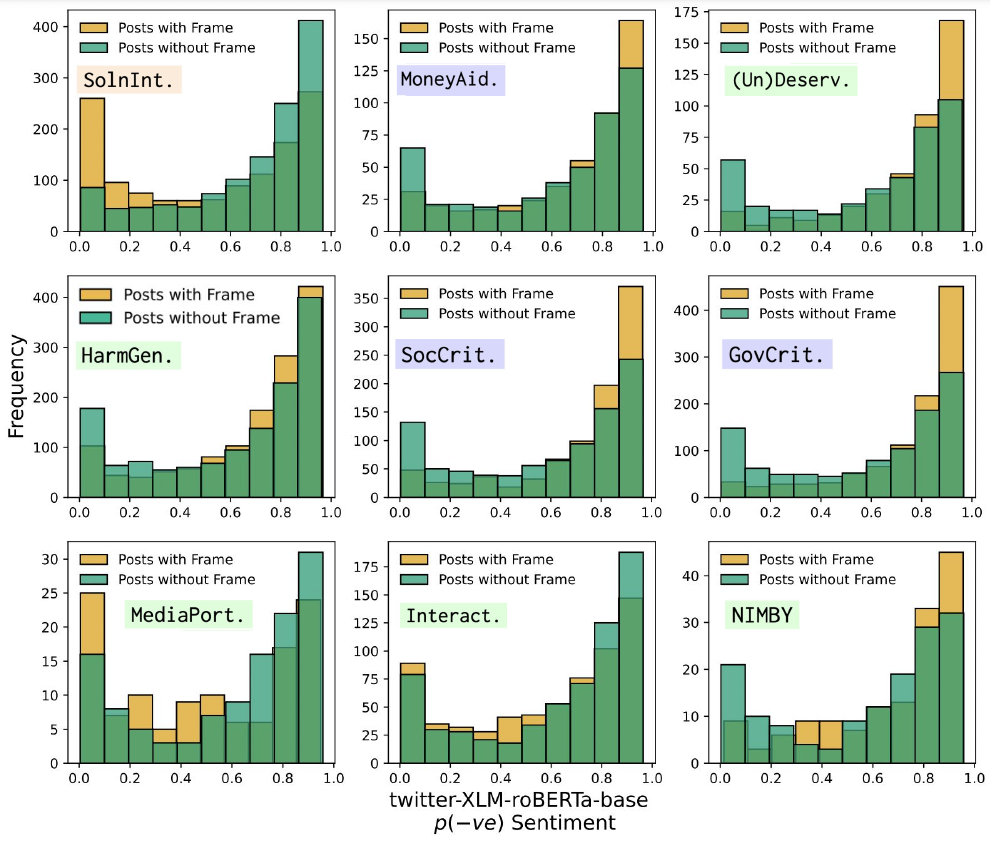}
 \caption{Distribution of twitter-XLM-roBERTa-base $p(-ve)$ sentiment probabilities \cite{rosenthal2017semeval} for posts labeled with each \finer frame in our subset of 4.1k human-annotated posts.}
 \label{fig:sent}
\end{figure*}

We include additional results from analyzing posts with \frames including predominant attitudes by state mentions in \autoref{fig:all_states} and corresponding examples of posts about states and their most salient attitudes in \autoref{tab:state_examples}. We also include the co-occurrence of PEH and vulnerable populations in \autoref{fig:co_occur_identity} and corresponding examples of the most significantly occurring attitudes and examples in posts containing co-occurrences of (PEH, Vulnerable Population) pairs in \autoref{tab:co_occur_identity} where we find marked differences amongst PEH when they are compared to Ukrainians vs. when they are compared to immigrants in posts that reflect \deservingundeserving. 
Our complete set of log-odds results from our analysis on \solutionsinterventions in CA is detailed in \autoref{tab:logodds} where we compute the log-odds ratio metric across CA posts comparing posts labeled with (+) \solutionsinterventions and not labeled with (-) \solutionsinterventions. We also compute the log-odds ratio metric between posts labeled with positive sentiment $p(+ve)$ and $p(-ve)$. 
Lastly, we provide a complete set of toxicity and sentiment distributions across all our \finer frames in \autoref{fig:tox} and \autoref{fig:sent}. 
 
\begin{table}[ht]
\centering
\resizebox{\linewidth}{!}{
\begin{tabular}{p{0.28\linewidth} p{0.99\linewidth}}
    \toprule
    \textbf{(State, Attitude)} & \textbf{Post}\\ 
    \midrule
    (New York, \moneyaid) & {\fontfamily{cmss}\selectfont  how much,money is allotted to illegal migrants? do you know? they sure are trashing out new york. everywhere they congregate there's piles of trash. how dare them through food away when we have so many homeless americans! funds need to help our people first!} \\
     \midrule
    (New York, \deservingundeserving) & {\fontfamily{cmss}\selectfont you’re right.  we should be feeding and housing our homeless veterans with our tax dollars instead of housing ungrateful illiberal immigrants in nyc high rises.  also - pay your own debt.} \\
    \midrule
    (California, \societalcritique) & {\fontfamily{cmss}\selectfont you idiots , help our homeless people first , help people in california that can’t even find food or clothes.} \\
     \midrule
    (California, \harmfulgeneralization) & {\fontfamily{cmss}\selectfont talk about how california has been ruined under you! los angeles and san francisco used to be wonderful, safe, clean cities! now they’re packed with homeless, drug addicts, violent lunatics, car thieves, murderers, and more. you’re a vile pig!} \\
   \midrule
    (Florida, \mediaportrayal) & {\fontfamily{cmss}\selectfont google alexia son peter. he absolutely beat the shit out of a homeless man on season 3. he’s also beat up his girlfriend. also he never even went to university of miami and is not allowed to be there.} \\
     \midrule
    (Oregon, \harmfulgeneralization) & {\fontfamily{cmss}\selectfont portland street response is a joke, i’m sorry.  the majority of the homeless who come to portland aren’t from here and likely to be criminals. have you missed the amount of fentanyl being dealt and the amount of guns within the homeless population?} \\
    \bottomrule 
\end{tabular}
}
\caption{\textbf{Examples of posts containing mentions of states and their corresponding salient attitudes:} Based on Figure \ref{fig:all_states}, we provide example posts for the most salient attitudes across posts containing state mentions.}
\label{tab:state_examples}
\end{table}


\begin{table*}[ht]
\centering
\resizebox{\linewidth}{!}{
\begin{tabular}{p{0.05\linewidth} p{0.12\linewidth}p{0.2\linewidth}p{0.99\linewidth}}
    \toprule
     & \textbf{Vulnerable Population}& \textbf{Significant \frames}& \textbf{Example}\\ 
    \midrule
    \multirow{2}{*}{\multirowcell{6}{\rotatebox[]{90}{`Homeless' Veteran}}}& \multicolumn{1}{>{\arraybackslash}p{0.12\linewidth}}{Ukrainian} & \multicolumn{1}{>{\arraybackslash}p{0.2\linewidth}}{\governmentcritique, \moneyaid}& \multicolumn{1}{>{\arraybackslash}p{0.99\linewidth}}{{\fontfamily{cmss}\selectfont @mention they have been given not 1 billon dollars, not 80 billion dollars (russian military budgets, but 180 billion dollars. do you really think that money is being used to help the ukrainian people-or is it something else? go help a homeless vet today!} \newline \textbf{Post labeled with: }\moneyaid, \governmentcritique, \deservingundeserving} \\\cline{2-4}
     & \multicolumn{1}{>{\arraybackslash}p{0.12\linewidth}}{Immigrant} & \multicolumn{1}{>{\arraybackslash}p{0.2\linewidth}}{\nimby}& \multicolumn{1}{>{\arraybackslash}p{0.99\linewidth}}{{\fontfamily{cmss}\selectfont@mention oh please they are dropping off bus loads of immigrants to hotels 6 blocks from my house that we the tax payers get to pay for. we have homeless vets what about them !! wake up} \newline \textbf{Post labeled with: }\nimby, \governmentcritique, \deservingundeserving} \\ \midrule 
            
    \multirow{2}{*}{\multirowcell{7}{\rotatebox[]{90}{`Homeless' People}}}& \multicolumn{1}{>{\arraybackslash}p{0.12\linewidth}}{ Ukrainian} & \multicolumn{1}{>{\arraybackslash}p{0.2\linewidth}}{\governmentcritique, \moneyaid}& \multicolumn{1}{>{\arraybackslash}p{0.99\linewidth}}{{\fontfamily{cmss}\selectfont ukrainian refugees can get ssi, housing benefits, and free health care. meanwhile, we have a lot of homeless people who have no access to such benefits - we have the highest number of the unemployed black population.} \newline \textbf{Post labeled with: }\moneyaid, \governmentcritique, \deservingundeserving} \\\cline{2-4}
    & \multicolumn{1}{>{\arraybackslash}p{0.12\linewidth}}{ Immigrant} & \multicolumn{1}{>{\arraybackslash}p{0.2\linewidth}}{\personalinteraction, \nimby, \harmfulgeneralization}& \multicolumn{1}{>{\arraybackslash}p{0.99\linewidth}}{{\fontfamily{cmss}\selectfont @mention new york city’s pulling the rug out from underneath the illegal immigrants kicking them out of the \$700 a day hotel rooms. was it the bedbugs or the thousands of empty bottles of beer and liquor. we used to have a problem with homeless people now we have a catastrophe of homeless} \newline \textbf{Post labeled with: }\nimby, \governmentcritique, \deservingundeserving} \\ \midrule

\multirow{2}{*}{\multirowcell{8}{\rotatebox[]{90}{`Homeless' American}}}& \multicolumn{1}{>{\arraybackslash}p{0.12\linewidth}}{ Ukrainian} & \multicolumn{1}{>{\arraybackslash}p{0.2\linewidth}}{ \governmentcritique, \moneyaid}& \multicolumn{1}{>{\arraybackslash}p{0.99\linewidth}}{{\fontfamily{cmss}\selectfont @mention grifter journalist got to grift . how about reporting on how close we are from being in a nuclearwar with russia. or how corrupted our politicians have become. write about how homeless americans live across this country and the us thinks sending money to ukraine is saving anyone} \newline \textbf{Post labeled with: }\moneyaid, \governmentcritique, \deservingundeserving} \\\cline{2-4}
& \multicolumn{1}{>{\arraybackslash}p{0.12\linewidth}}{Immigrant} & \multicolumn{1}{>{\arraybackslash}p{0.2\linewidth}}{\societalcritique, \solutionsinterventions, \personalinteraction, \mediaportrayal, \nimby, \harmfulgeneralization}&\multicolumn{1}{>{\arraybackslash}p{0.99\linewidth}}{{\fontfamily{cmss}\selectfont @mention help homeless americans that are being discriminated against by the huge number of foreign aliens in america or give money to another county. i say americans are f***ed until we start saying no to everything immigrant. more foreign aliens = more poor americans = communism grows} \newline \textbf{Post labeled with: }\societalcritique, \solutionsinterventions, \deservingundeserving} \\
    \bottomrule 
\end{tabular}
}
\caption{\textbf{Statistically significant frames for posts containing most frequent (PEH, Vulnerable Population) co-occurrence pairs.} For each subset of posts containing a (PEH, Vulnerable Population) co-occurrence pair, we find the most statistically significant frames ($p < 0.0001$). We observe that posts containing comparisons between PEH and Ukrainians elicit more \critiques whereas posts comparing PEH to immigrants elicit more \nimby and \harmfulgeneralization attitudes. }
\label{tab:co_occur_identity}
\end{table*}

\begin{table*}[ht!]
\centering
\small
\resizebox{0.98\linewidth}{!}{
\begin{tabular}{ccccccccc}
    \toprule
    \textbf{\finer} & \textbf{Metrics} & {\textbf{\texttt{Experts}}} & {\textbf{\gptsolo}} & {\textbf{\gptintheloop}} & {\textbf{\flansmall}} & {\textbf{\flanbase}} &{\textbf{\flanlarge}} & {\textbf{\flanxl}}\\ 
    \midrule
    \multirow{3}{*}\governmentcritique & {Prec.} & $0.69 \pm 0.17$ & $0.64 \pm 0.16$ & $0.68 \pm 0.16$ & $0.59 \pm 0.15$ & $0.66 \pm 0.14$ & $0.61 \pm 0.13$ & $0.65 \pm 0.14$ \\
     & Rec. & $0.69 \pm 0.17$ & $0.82 \pm 0.09$ & $0.81 \pm 0.09$ & $0.79 \pm 0.07$ & $0.67 \pm 0.09$ & $0.74 \pm 0.09$ & $0.70 \pm 0.10$\\
     & F1 & $0.65 \pm 0.06$ & $0.7 \pm 0.08$ & $0.72 \pm 0.07$ &$0.66 \pm 0.09$  & $0.65 \pm 0.06$ & $0.65 \pm 0.07$ &  $0.65 \pm 0.05$\\
     \midrule
    \multirow{3}{*}\moneyaid & Prec. & $0.66 \pm 0.22$ & $0.45 \pm 0.14$ & $0.5 \pm 0.16$ & $0.66 \pm 0.16$ & $0.75 \pm 0.16$ & $0.60 \pm 0.23$ & $0.54 \pm 0.17$\\
     & Rec. & $0.66 \pm 0.22$ & $0.87 \pm 0.06$ & $0.86 \pm 0.07$ & $0.70 \pm 0.17$ & $0.72 \pm 0.16$ & $0.61 \pm 0.07$ & $0.85 \pm 0.09$\\
     & F1 & $0.60 \pm 0.16$ & $0.58 \pm 0.14$ & $0.61 \pm 0.15$ & $0.65 \pm 0.10$ & $0.70 \pm 0.09$ & $0.59 \pm 0.18$ & $0.63 \pm 0.15$\\
     \midrule
         \multirow{3}{*}\societalcritique & Prec. & $0.46 \pm 0.15$ & $0.36 \pm 0.09$ & $0.47 \pm 0.09$ & $0.28 \pm 0.05$ & $0.35 \pm 0.11$ & $0.44 \pm 0.15$ & $0.50 \pm 0.15$\\
     & Rec. & $0.46 \pm 0.15$ & $0.71 \pm 0.12$ & $0.56 \pm 0.13$ & $0.36 \pm 0.12$ & $0.43 \pm 0.15$ & $0.43 \pm 0.11$ & $0.42 \pm 0.13$\\
     & F1 & $0.43 \pm 0.09$ & $0.46 \pm 0.06$ & $0.5 \pm 0.05$ & $0.31 \pm 0.07$ & $0.37 \pm 0.10$ & $0.42 \pm 0.09$ & $0.44 \pm 0.10$\\
     \midrule
         \multirow{3}{*}\solutionsinterventions & Prec. & $0.70 \pm 0.11$ & $0.54 \pm 0.07$ & $0.65 \pm 0.07$ & $0.53 \pm 0.06$ & $0.54 \pm 0.07$ & $0.67 \pm 0.04$ & $0.59 \pm 0.06$\\
     & Rec. & $0.70 \pm 0.11$ & $0.72 \pm 0.09$ & $0.66 \pm 0.1$ & $0.66 \pm 0.02$ & $0.69 \pm 0.05$ & $0.70 \pm 0.06$ & $0.72 \pm 0.03$ \\
     & F1 & $0.69 \pm 0.06$ & $0.62 \pm 0.06$ & $0.65 \pm 0.06$ & $0.59 \pm 0.04$ & $0.60 \pm 0.03$ & $0.68 \pm 0.03$ & $0.64 \pm 0.03$\\
     \midrule
         \multirow{3}{*}\personalinteraction & Prec. & $0.48 \pm 0.18$ & $0.34 \pm 0.07$ & $0.45 \pm 0.1$ & $0.54 \pm 0.10$ & $0.41 \pm 0.11$ & $0.53 \pm 0.11$ & $0.46 \pm 0.13$\\
     & Rec. & $0.48 \pm 0.18$ & $0.73 \pm 0.13$ & $0.66 \pm 0.15$ & $0.34 \pm 0.12$ & $0.57 \pm 0.13$ & $0.56 \pm 0.14$ & $0.58 \pm 0.22$\\
     & F1 & $0.45 \pm 0.13$ & $0.45 \pm 0.07$ & $0.52 \pm 0.07$ & $0.41 \pm 0.10$ & $0.46 \pm 0.09$ & $0.54 \pm 0.11$ & $0.50 \pm 0.15$\\
     \midrule
         \multirow{3}{*}\mediaportrayal & Prec. & $0.73 \pm 0.18$ & $0.41 \pm 0.09$ & $0.6 \pm 0.11$ & $0.67 \pm 0.47$ & $0.47 \pm 0.14$ & $0.75 \pm 0.16$ & $0.42 \pm 0.33$ \\
     & Rec. & $0.73 \pm 0.18$ & $0.69 \pm 0.14$ & $0.67 \pm 0.15$ & $0.17 \pm 0.13$ & $0.57 \pm 0.10$ & $0.64 \pm 0.08$ & $0.35 \pm 0.26$\\
     & F1 & $0.70 \pm 0.10$ & $0.5 \pm 0.08$ & $0.63 \pm 0.11$ & $0.26 \pm 0.20$ & $0.51 \pm 0.11$ & $0.69 \pm 0.10$ & $0.37 \pm 0.28$\\
     \midrule
     \multirow{3}{*}\nimby & Prec. & $0.20 \pm 0.26$ & $0.27 \pm 0.19$ & $0.48 \pm 0.39$ & $0.21 \pm 0.37$ & $0.30 \pm 0.22$ & $0.50 \pm 0.42$ & $0.42 \pm 0.35$ \\
     & Rec. & $0.20 \pm 0.26$ & $0.47 \pm 0.29$ & $0.47 \pm 0.29$ & $0.05 \pm 0.07$ & $0.21 \pm 0.18$ & $0.28 \pm 0.21$ & $0.61 \pm 0.34$\\
     & F1 & $0.17 \pm 0.21$ & $0.3 \pm 0.16$ & $0.39 \pm 0.21$ & $0.08 \pm 0.12$ & $0.23 \pm 0.17$ & $0.34 \pm 0.26$ & $0.45 \pm 0.32$\\
     \midrule
     \multirow{3}{*}\deservingundeserving & Prec. & $0.61 \pm 0.22$ & $0.32 \pm 0.06$ & $0.56 \pm 0.07$ & $0.57 \pm 0.25$ & $0.64 \pm 0.17$ & $0.64 \pm 0.14$ & $0.60 \pm 0.15$ \\
     & Rec. & $0.61 \pm 0.22$ & $0.87 \pm 0.11$ & $0.75 \pm 0.17$ & $0.33 \pm 0.09$& $0.59 \pm 0.13$ & $0.73 \pm 0.17$ & $0.73 \pm 0.17$\\
     & F1 & $0.58 \pm 0.19$ & $0.46 \pm 0.06$ & $0.63 \pm 0.08$ & $0.41 \pm 0.13$ & $0.60 \pm 0.11$ & $0.67 \pm 0.10$ & $0.64 \pm 0.11$ \\
     \midrule
     \multirow{3}{*}\harmfulgeneralization & Prec. & $0.54 \pm 0.11$ & $0.53 \pm 0.13$ & $0.58 \pm 0.15$ & $0.50 \pm 0.12$ & $0.59 \pm 0.19$ & $0.61 \pm 0.13$ & $0.57 \pm 0.15$\\
     & Rec. & $0.54 \pm 0.11$ & $0.48 \pm 0.07$ & $0.48 \pm 0.07$ & $0.49 \pm 0.10$ & $0.30 \pm 0.05$ & $0.51 \pm 0.05$ & $0.58 \pm 0.08$ \\
     & F1 & $0.53 \pm 0.07$ & $0.49 \pm 0.06$ & $0.51 \pm 0.06$ & $0.48 \pm 0.06$ & $0.39 \pm 0.07$ & $0.55 \pm 0.05$ & $0.56 \pm 0.05$\\
     \midrule
     \multirow{3}{*}\filter & Prec. & $0.76 \pm 0.10$ & $0.14 \pm 0.2$ & $0.14 \pm 0.2$ & $0.00 \pm 0.00$ & $0.00 \pm 0.00$ & $0.00 \pm 0.00$ & $0.00 \pm 0.00$\\
     & Rec. & $0.76 \pm 0.07$ & $0.02 \pm 0.03$ & $0.02 \pm 0.03$ & $0.00 \pm 0.00$ & $0.00 \pm 0.00$ & $0.00 \pm 0.00$ & $0.00 \pm 0.00$ \\
     & F1 & $0.75 \pm 0.02$ & $0.04 \pm 0.05$ & $0.02 \pm 0.03$ & $0.00 \pm 0.00$ & $0.00 \pm 0.00$ & $0.00 \pm 0.00$ & $0.00 \pm 0.00$\\
    \midrule
     \multirow{4}{*}{Aggregated ($|D_\text{agreement}|=$280)}& {Prec.} & 0.58 $\pm$ 0.08 & 0.39 $\pm$ 0.05 & 0.51 $\pm$ 0.07 & 0.45 $\pm$ 0.08  & 0.47 $\pm$ 0.06 & 0.54 $\pm$ 0.07 & 0.47 $\pm$ 0.06\\
    & Rec. & 0.58 $\pm$ 0.08 & 0.64 $\pm$ 0.05  & 0.6 $\pm$ 0.05  & 0.39 $\pm$ 0.03 & 0.47 $\pm$ 0.05 & 0.52 $\pm$ 0.05 & 0.55 $\pm$ 0.03 \\
    & F$_1$ & 0.55 $\pm$ 0.04 & 0.46 $\pm$ 0.04 & 0.52 $\pm$ 0.04 & 0.38 $\pm$ 0.04 & 0.45 $\pm$ 0.04 & 0.51 $\pm$ 0.03 &  0.49 $\pm$ 0.03 \\
    &  Fleiss' $\kappa$ & 0.46 $\pm$ 0.04 & 0.31 $\pm$ 0.04  & 0.37 $\pm$ 0.04 & 0.24 $\pm$ 0.01 & 0.31 $\pm$ 0.02 &  0.34 $\pm$ 0.03 & 0.38 $\pm$ 0.04\\
    \midrule
     \multirow{3}{*}{Aggregated ($|D_\text{test}|=$1280)}& {Prec.} & - & - & - & 0.46 & 0.48 & 0.50 & 0.47 \\
    & Rec. & - & - & - & 0.41 & 0.47 & 0.48 & 0.51 \\
    & F$_1$ & - & - & - & 0.42 & 0.47 & 0.50 & 0.48 \\
    \bottomrule 
\end{tabular}
}

\caption{Precision, Recall and F$_1$ for each of the \finer frames and overall macro F$_1$ (Aggregated) on ($|D_\text{agreement}|=$280). Due to compute restrictions, we do not generate \gptsolo and \gptintheloop predictions on ($|D_\text{test}|=$1280).}
\label{tab:res:finer_models_vs_everyone}
\end{table*}

\begin{table*}[h!]
\centering
\footnotesize
\resizebox{\textwidth}{!}{
\begin{tabular}{lclclclc}
\toprule
(+) \solutionsinterventions & Score & (-) \solutionsinterventions & Score & $p(+ve)$ & Score & $p(-ve)$ & Score \\
\midrule
rent & 3.72 & crime & -7.30 & great & 4.68 & drug & -3.84\\
voucher & 2.59 & drug & -6.73 & love & 4.22 & crime & -2.74\\
abandon & 2.48 & illegal & -5.76 & new & 3.26 & addict & -2.61\\
bass & 4.68 & shit & -5.62 & thank & 3.16 & democrat & -2.24\\
healthcare & 2.03 & tax & -4.73 & help & 2.32 & highest & -2.12\\

\bottomrule
\end{tabular}
}
\caption{Log-odds ratios for terms that are statistically more likely to appear in posts mentioning California and labeled with (+) \solutionsinterventions, posts not labeled with (-) \solutionsinterventions, posts with positive sentiment ($p(+ve) > 0.5$) and negative sentiment ($p(-ve) > 0.5$). Scores outside the range [1.96, -1.96] are considered statistically significant. 
}
\label{tab:logodds}
\vspace{-0.2cm}
\end{table*}

\clearpage 

\section{Further Related Work}
\label{sec-related}

\subsection{Characterizing Attitudes Towards Homelessness}
\label{sec:related-attitudes}
Prior work characterizes public opinion and attitudes towards homelessness through ethnographic studies, surveys, content analyses and psychometric analyses on small samples of human participants \cite{gowan2010hobos, survey_highschool_1997, attitudes_general_public_survey199, attitudes_robert_2015, kingree1997preliminary}. 
Prior studies have found that diversity in public attitudes towards homelessness are heavily influenced by personal experiences and socio-demographic backgrounds. More specifically, \citet{tsai2019public} conducted a survey of 541 U.S. adults on public attitudes about homelessness with the goal of determining  socio-demographic characteristics of survey respondents that are associated with particular attitudes. They find that a majority of the participants who identified as female, lower income, Democrat and were personally exposed to homelessness expressed attitudes of compassion towards PEH and believed that the federal government should take more constructive measures in funding and reform policies for PEH. Similarly, \citet{smith2023attitudesincome} and \citet{mullenbach2023socialcontact} study people's prior personal experiences with homelessness and its effect on attitudes and perceptions towards PEH. 
Prior studies also analyze how sociodemographic backgrounds of participants influence an individual's belief behind the cause of homelessness through surveys \cite{lee1990causes, Dittmeier2018PerceptionsOH, attitudes_robert_2015, hockingprosocial2000}. Such studies on attitudes and perceptions towards PEH play an important role in influencing public policies that address homelessness and restrictive housing policies \cite{mullenbach2023socialcontact, compassionfatigue1995}. 
Public opinion is heavily influenced by the social and political climate. To this end, prior work have also looked at changing attitudes towards homelessness from 1990 to 2016 \cite{changes_in_attitudes2017} and 1993 to 2001 \cite{changes_in_attitudes2006} by comparing survey responses conducted in different years and they found increasingly more liberal and compassionate responses towards homelessness.
Beyond analyzing survey responses, \citet{inbook} identifies a basic topology from ethnographic studies to identify key narratives in discourse about homelessness: sin-talk, sick-talk, and system-talk which characterize the causes of homelessness enlisting criminal behavior and mental illness as primary contributing factors. Although such studies provide invaluable insights into public opinion, it is extremely challenging to collect and analyze survey responses at a larger scale. Furthermore, prior work has also identified shortcomings in existing datasets and toxicity classifiers in recognizing degrading rhetoric against poor and disadvantaged populations. This is because harmful language against marginalized groups is often masked by humor and sarcasm and includes more subtle accounts of stereotyping that may be specific to a particular vulnerable population and thus harder to recognize \cite{kiritchenko-etal-2023-aporophobia, curto2023crime}. As a result, there exist few specialized datasets to study language targeted specifically towards PEH \cite{kiritchenko-etal-2023-aporophobia}. 
To this end, we plan to expand on these studies by performing large-scale analyses on social media posts annotated with \frames and providing a data resource that analyzes complex, and at times, harmful attitudes against PEH. 

\subsection{Quantifying Framing in Social Media Discourse.}
\label{sec:related-quantifying-framing}

Framing \cite{framing_entman} offers one mechanism to understand the communicative power of language by characterizing what people find most important in a text. Framing has been used to analyze the politics of homelessness and crime in New York City to characterize responses to homelessness \cite{Upton_2016}. Furthermore, framing has also been used to study discourse on social media for topics such as immigration and politics \cite{mendelsohn2021modeling, Boydstun2013IdentifyingMF}. Our line of work is most similar to prior work that studies stigma and stereotypes against PEH in the media \cite{coco_berkeley_homelessness, imageslink1991, leibowitz2005stereotypes, belcher2012stigma}. 
More specifically, \citet{coco_berkeley_homelessness} characterize stigma against PEH using social media posts by manually annotating 6.4k posts and defining a set of codes that characterize different stigmatizing traits attributed to PEH. We build on prior work that use framing theory to surface \finer frames that characterize attitudes towards homelessness on social media and apply \frames to study attitudes at scale. 

\subsection{LLMs as an Annotation Assistant.}
\label{sec:related-llm-assistant}

Collecting high quality annotations for socially complex, and sensitive tasks is not only a laborious and time consuming process, but also mentally taxing. Prior work have explored using LLMs as an assistant in the annotation process to resolve ambiguities and reduce the burden off of human annotators \cite{li2023coannotating, rytting2023towards}. 
However, \citet{pangakis2023automated} test the performance of GPT-4 across 27 annotation tasks and find that LLM performance varies widely across tasks and is largely dependent on the difficulty and subjectivity of the annotation task. 
They suggest evaluating generated annotations with a set of high quality expert annotations during prompt tuning. 
Furthermore, \citet{rytting2023towards} experiment with GPT-3 as an annotator on its own and find that across four different tasks, GPT-3 performs more efficiently and as effectively as human annotators. 
For socially complex tasks such as homelessness, we use \gptsolo as an assistant in the annotation process while validating its predictions with human experts to leverage both the efficiency while also maintaining of quality of LLM generated annotations. It is important for us to validate \gptsolo predictions as LLMs have been shown to misrepresent identity groups which can cause downstream harms when reasoning about problems with sensitive social implications such as homelessness \cite{Wang2024LargeLM}.

\end{document}